\documentclass[acmtog]{acmart}
\usepackage{subcaption}
\usepackage{float}
\usepackage{makecell}
\usepackage{booktabs} %
\usepackage{tabularx} %
\newcolumntype{Y}{>{\centering\arraybackslash}X} %
\usepackage{graphicx} %
\usepackage{amsmath}  %
\usepackage{tikz}
\usepackage{placeins} %

\usetikzlibrary{positioning,calc,shapes.arrows}
\AtBeginDocument{%
  }

\copyrightyear{2026}
\acmYear{2026}
\setcopyright{cc}
\setcctype{by-nc-nd}
\acmConference[SA Conference Papers '26]{SIGGRAPH Asia 2026 Conference Papers}{December 01--04, 2026}{Kuala Lumpur, Malaysia}
\acmBooktitle{SIGGRAPH Asia 2026 Conference Papers (SA Conference Papers '26), December 01--04, 2026, Kuala Lumpur, Malaysia}
\acmDOI{10.1145/3829340.3842358}
\acmISBN{979-8-4007-2842-6/2026/12}

\begin{document}

\title{FaceRiff: Data-Efficient Unified Text-to-3D Face Asset Synthesis via Geometric Conditioning}

\author{Elly Akhoundi}
\affiliation{%
  \institution{Electronic Arts}
  \city{Vancouver}
  \country{Canada}}
\email{eakhoundi@ea.com}

\author{Mathieu Lamarre}
\affiliation{%
  \institution{Electronic Arts}
  \city{Montréal}
  \state{Québec}
  \country{Canada}
}
\email{mlamarre@ea.com}

\author{Thibault Goyallon}
\affiliation{%
  \institution{Electronic Arts}
  \city{Toronto}
  \state{Ontario}
  \country{Canada}
}
\email{tgoyallon@ea.com}

\author{Patrick Anderson}
\affiliation{%
  \institution{Electronic Arts}
  \city{Montréal}
  \state{Québec}
  \country{Canada}
}
\email{PAnderson@ea.com}

\author{Étienne Danvoye}
\affiliation{%
  \institution{Electronic Arts}
  \city{Montréal}
  \state{Québec}
  \country{Canada}
}
\email{edanvoye@ea.com}

\renewcommand{\shortauthors}{Akhoundi et al.}  

\begin{abstract}
High-resolution facial datasets with consistent UV-mapped geometry, albedo, and physically based displacement are scarce, yet such assets are essential for photorealistic digital-human rendering. Existing 3D face generation methods are often slow, single-modality, or require large specialized datasets. Meanwhile, large image diffusion models encode strong facial priors but do not natively produce lighting-independent textures, pore-scale displacement, or UV-aligned geometry. We present \textbf{FaceRiff}, a unified, data- and parameter-efficient framework for generating complete 3D face assets (mesh, albedo, and displacement) directly from text in a shared UV space. FaceRiff adapts a pretrained SDXL backbone with task-specific LoRA adapters and uses UV normal maps as geometric conditioning: per-instance normals guide albedo and displacement synthesis, while a dataset-mean normal provides a canonical spatial anchor for mesh generation. Using only $2{,}049$ scans from approximately 350 identities and training on a single GPU, FaceRiff updates less than $3\%$ of the backbone parameters per task and produces render-ready 3D face assets without per-instance optimization. Compared with FaceG2E and DreamFace, FaceRiff produces greater per-prompt diversity while preserving prompt alignment and fixed-topology UV consistency, and additionally generates lighting-independent albedo suitable for PBR relighting and pore-scale displacement absent from prior text-driven baselines. A blind study with $45$ industry professionals shows that FaceRiff achieves substantially higher perceived realism than DreamFace, approaching that of real captures. Overall, FaceRiff demonstrates that pretrained diffusion priors can be adapted to unified 3D face asset synthesis under limited data.
\end{abstract}

\ccsdesc[500]{Computing methodologies~Animation}
\ccsdesc[500]{Computing methodologies~Computer vision}
\ccsdesc[300]{Computing methodologies~Shape modeling}
\ccsdesc[300]{Computing methodologies~Neural networks}

\keywords{Digital Humans, 3D Face Synthesis, Diffusion Models, PBR Texture Synthesis, Text-to-3D, UV Position Maps, ControlNet, Parameter-Efficient Fine-Tuning, VLM-based Tagging}

\begin{teaserfigure}
    \centering
    \includegraphics[width=0.8\textwidth]{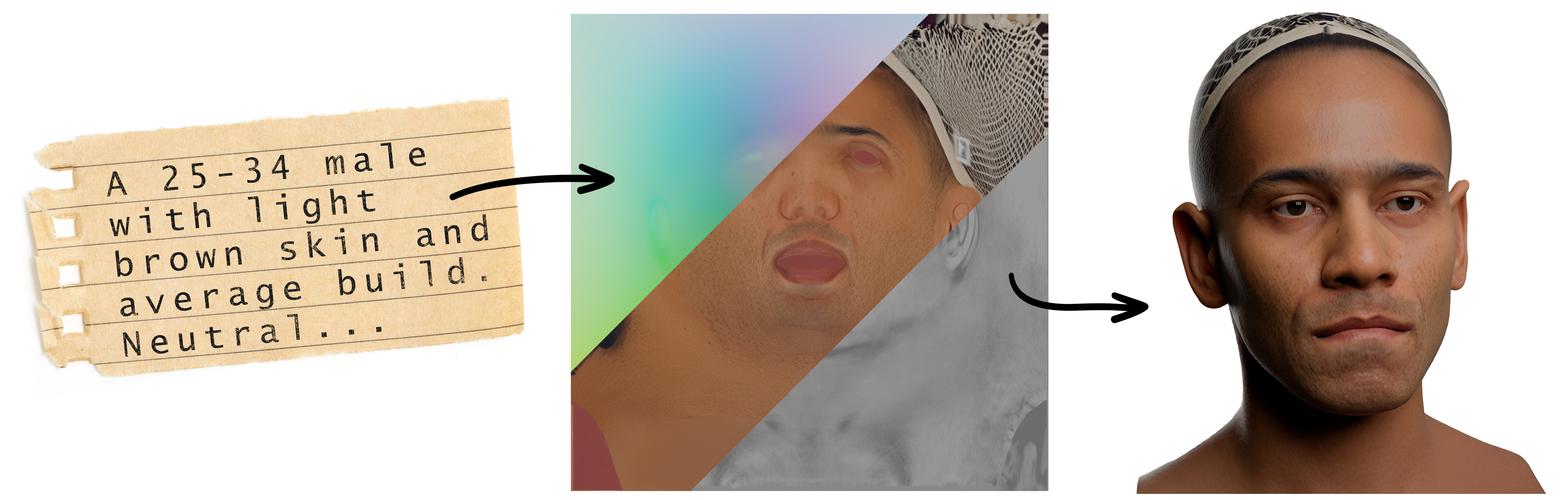}
    \caption{\textbf{FaceRiff: text-driven 3D face asset generation.} From a text prompt, FaceRiff synthesizes mesh, lighting-independent albedo, and pore-scale displacement in a shared UV space in a feed-forward manner, combining them into the final rendered full head.}
    \Description{Teaser figure showing a text prompt alongside the generated 3D face assets---mesh, albedo, displacement---and the final rendered head, produced feed-forward by FaceRiff.}
    \label{fig:banner}
\end{teaserfigure}

\maketitle

\section{Introduction}
Photorealistic digital humans depend on a coupled set of assets: an accurate mesh, lighting-independent albedo, and physically based displacement capturing pore- and wrinkle-scale geometry. Acquiring these assets requires costly multi-view capture systems and extensive processing, limiting access for independent creators and slowing iteration.

Existing generative approaches address only part of this problem. Score-distillation methods~\cite{Zhang2023DreamFace,wu2023faceg2e,liu2024headartist,huang2024humannorm} produce high-quality results but are slow and may yield inconsistent topology. UV-space latent diffusion models trained from scratch~\cite{otto2024multimodal} recover topology-consistent geometry but require large UV datasets and produce only geometry. Biophysical and parametric methods~\cite{liu2025biophysical} deliver PBR-relightable faces but rely on auxiliary simulation and offer no free-form text interface. Other approaches target albedo~\cite{yang2024self,ran2024hifi} or head meshes~\cite{zielonka2022mica} in isolation. Recent work adapts pretrained image diffusion priors for 3D-face tasks such as reflectance estimation~\cite{ren2025s3face}, UV texture recovery~\cite{yang2025freeuv}, animatable avatars~\cite{anim3davatar2025}, and full-body avatars via SDXL adapters~\cite{dream3davatar2025}, showing that image priors transfer to 3D-related signals. However, existing methods do not generate a complete UV-aligned mesh, albedo, and displacement asset stack directly from text. Compounding this gap, high-resolution UV-mapped facial assets remain scarce~\cite{Bai_2023_CVPR,dib2024mosar}, making data-hungry approaches impractical.

We present FaceRiff, a data- and parameter-efficient framework that adapts a pretrained SDXL backbone~\cite{podell2024sdxl} with task-specific LoRA adapters~\cite{hu2022lora} to generate a coherent mesh, albedo, and displacement asset stack directly from text in a shared UV space. Our key observation is that, in the low-data UV-face setting, UV-space geometry provides an effective spatial prior for adapting pretrained image diffusion models to aligned 3D facial asset generation. The contribution is in the system and representation design, not a new diffusion architecture.

Building on this observation, FaceRiff combines three design choices for unified text-to-3D face asset synthesis under limited data. \textbf{(i) Normal-map geometric conditioning.} For albedo and displacement, per-instance UV normals guide synthesis toward anatomically appropriate UV regions; for mesh, before per-instance geometry is available, a dataset-mean UV normal provides a canonical anchor. \textbf{(ii) Residual UV position maps for mesh synthesis.} We predict per-identity offsets from a dataset-mean UV position map, using percentile-based normalization to map geometry to the pretrained VAE range while preserving high-frequency detail. \textbf{(iii) Rich semantic captions.} A vision-language model annotates each scan with structured descriptions of appearance, facial geometry, skin detail, and FACS-grounded~\cite{ekman2002facs} expression cues.

Our contributions are:
\begin{itemize}
    \item \textbf{A unified text-to-3D face asset framework} that generates a render-ready mesh, albedo, and displacement stack in a shared UV space with UV-consistent cross-modality alignment without multi-task losses or post-hoc registration.
    \item \textbf{A geometric conditioning strategy for low-data UV facial asset synthesis} that uses UV normals as spatial guidance across the generated asset stack.
    \item \textbf{A mean-anchored residual UV position-map formulation for text-driven mesh synthesis} that predicts offsets from a dataset-mean position map with percentile normalization, enabling fixed-topology geometry generation under limited data.
    \item \textbf{Data- and parameter-efficient training} with $2{,}049$ scans from approximately 350 identities, a single GPU, and LoRA adapters updating less than $3\%$ of the backbone.
\end{itemize}

\section{Related Work}
High-fidelity 3D facial asset synthesis has evolved from GAN-based UV texture priors~\cite{gecer2019ganfit,lattas2020avatarme} toward diffusion-based generative models~\cite{dhariwal2021diffusion,rombach2022high}, which offer strong priors for diverse pores, wrinkles, and identity-specific skin detail. However, adapting pretrained image diffusion models to facial UV assets remains challenging because the target maps represent different kinds of data: albedo maps store image-like surface color, while position and displacement maps store geometric measurements whose values do not follow natural-image statistics. We organize related work around four aspects of FaceRiff: text-driven 3D face generation, UV-space facial asset synthesis and reflectance, adaptation of pretrained image diffusion priors, and diffusion-based 3D texture generation. Related efforts also target text-guided head models~\cite{manu2023cliphead}, sphere-anchored face-geometry diffusion~\cite{zhang2026spherical}, ultra-high-resolution face textures~\cite{gruber2024gantlitz}, and diffusion-driven diversity for semantic-guided face assets~\cite{cai2026diversity}.

\subsection{Text-Driven 3D Face Generation}
Score-distillation methods~\cite{Zhang2023DreamFace,wu2023faceg2e,liu2024headartist,huang2024humannorm} optimize 3D representations against text-to-image diffusion priors, producing high-quality results but requiring minutes to hours per sample. DreamFace~\cite{Zhang2023DreamFace} and FaceG2E~\cite{wu2023faceg2e} preserve topology through parametric 3DMMs but constrain variation to the model subspace, while HeadArtist~\cite{liu2024headartist} and HumanNorm~\cite{huang2024humannorm} use freer representations with less consistent topology. Feed-forward tri-plane models such as Rodin~\cite{wang2023rodin} generate 3D heads directly, but do not naturally decompose outputs into mesh, albedo, and displacement. ClipFace~\cite{aneja2023clipface} edits StyleGAN UV textures for FLAME under text guidance, but remains tied to FLAME geometry and expression space.

UV-space latent diffusion models trained from scratch~\cite{otto2024multimodal} generate topology-consistent geometry from diverse conditioning signals in a single pass. We build on residual UV position maps, but differ in setting and scope: FaceRiff anchors residuals to a dataset-mean UV position map, adapts a pretrained SDXL backbone~\cite{podell2024sdxl} with LoRA modules~\cite{hu2022lora} using only $2{,}049$ scans rather than training from scratch on ${\sim}200$k UV maps, and extends UV-space diffusion from geometry-only generation to a shared mesh, albedo, and displacement asset stack.

Biophysical face generation offers another form of control: Liu et al.~\cite{liu2025biophysical} produce PBR-relightable faces with continuous control over melanin, hemoglobin, and age via auxiliary simulation, but use a fixed parametric interface without free-form text prompts. In contrast, FaceRiff uses VLM-derived dermatological, anatomical, and FACS-aware captions as a text-based semantic control interface.

\subsection{UV-Space Facial Asset Generation and Reflectance}
Prior work operating directly in facial UV space primarily targets reconstruction, reflectance estimation, or single asset modalities rather than text-driven asset synthesis. Relightify~\cite{papantoniou2023relightify} completes UV reflectance maps from partial observations; FitMe~\cite{lattas2023fitme} and AvatarMe++~\cite{lattas2021avatarme++} recover PBR-relightable UV assets from a single image using a StyleGAN prior or feed-forward image translation, building on single-image reflectance and geometry inference~\cite{saito2017photorealistic,yamaguchi2018highfidelity,li2020learning}. MoSAR~\cite{dib2024mosar} reconstructs translucency, ambient occlusion, and 4K displacement via differentiable shading, while other lines focus on albedo~\cite{yang2024self,ran2024hifi} or head-mesh reconstruction~\cite{zielonka2022mica}. In contrast, FaceRiff targets text-driven synthesis of a coherent mesh, albedo, and displacement asset stack in a shared UV space.

\subsection{Foundation-Model Adaptation for 3D-Face Tasks}
Recent work adapts pretrained image diffusion priors to 3D face tasks, including SSS-compliant reflectance estimation~\cite{ren2025s3face}, ground-truth-free UV texture recovery~\cite{yang2025freeuv}, animatable avatars conditioned on morphable models~\cite{anim3davatar2025}, and full-body avatars via SDXL adapters~\cite{dream3davatar2025}. These works motivate our approach but target only individual asset types or modalities. In contrast, FaceRiff studies geometric conditioning for unified UV-space generation of mesh, albedo, and displacement under limited data, using SDXL~\cite{podell2024sdxl} for its native resolution and text fidelity.

\subsection{Diffusion-Based 3D Texture Generation}
General-object texturing methods such as Text2Tex~\cite{chen2023text2tex} and TEXTure~\cite{richardson2023texture} project 2D diffusion samples onto meshes iteratively, while Paint3D~\cite{zeng2024paint3d} reduces baked illumination with lighting-less UV diffusion and MVPainter~\cite{shao2025mvpainter} improves view consistency through multi-view conditioning. Direct UV-space approaches further improve coherence, including Point-UV Diffusion~\cite{yu2023point}, TEXGen~\cite{yu2024texgen}, and UniTEX~\cite{liang2025unitex}. These methods target topological generalization across object categories, whereas FaceRiff exploits fixed facial UV topology to focus capacity on fine-scale pores, wrinkles, and expression lines. Geometry-Aware Texture Generation~\cite{Fadaeinejad2025GeometryAwareTG} is closer in domain, but uses a feed-forward GAN/VAE for neutral-expression head modeling without text prompts.

\section{Methodology}
\label{sec:method}

\begin{figure*}[t]
  \centering
  \includegraphics[width=\textwidth]{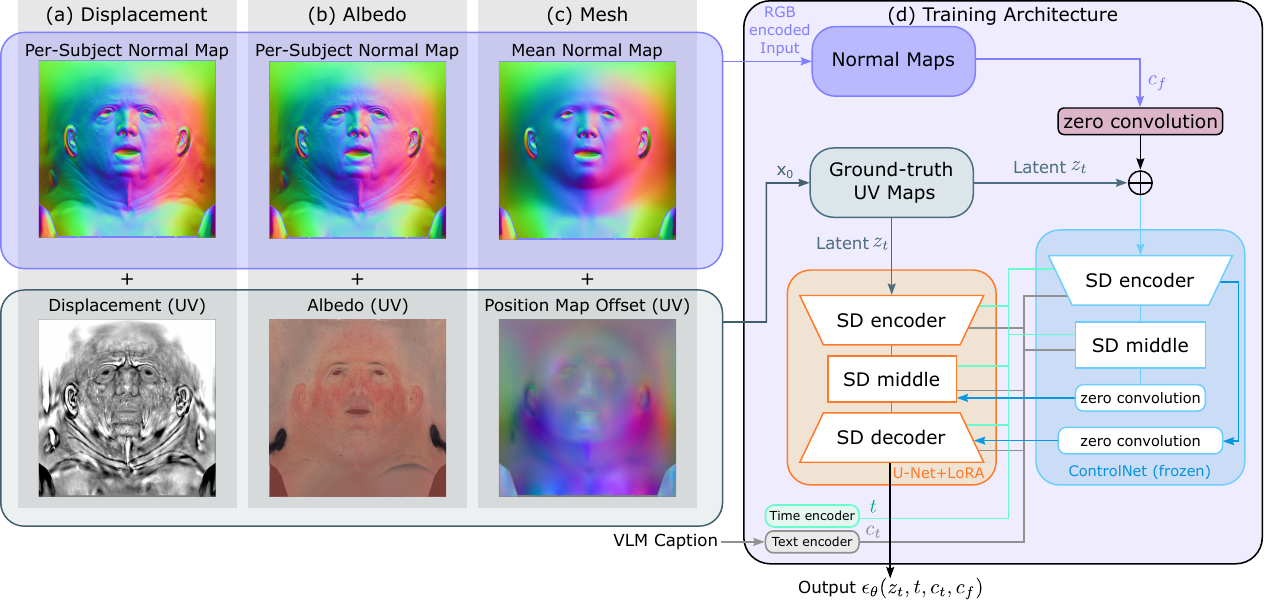}
    \vspace{-12pt}
    \caption{
    \textbf{FaceRiff pipeline overview.}
    \textbf{(a--c)} Task-specific LoRA adapters synthesize displacement, albedo, and mesh UV maps from a VLM caption and normal conditioning. Displacement and albedo use per-subject normals \(N\), while mesh generation uses the dataset-mean normal \(\bar{N}\) as a canonical scaffold.
    \textbf{(d)} The ground-truth UV map is encoded as \(z_0\), noised to \(z_t\), and denoised by SDXL to predict \(\epsilon_\theta(z_t, t, c_t, c_f)\). A frozen Normal ControlNet provides normal conditioning \(c_f\), while task-specific LoRA weights adapt SDXL for each modality.
    }
  \label{fig:overview}
\end{figure*}

FaceRiff generates mesh, albedo, and displacement in UV space with a pretrained SDXL backbone~\cite{podell2024sdxl}, task-specific LoRA adapters, and a frozen Normal ControlNet~\cite{zhang2023controlnet,xinsir2024controlnetunion}. All three tasks share the UV layout, captions, and adaptation recipe, but differ in conditioning geometry: mesh uses a dataset-mean normal $\bar{N}$ as a spatial scaffold, while albedo and displacement use per-instance normals $N$. Experimental details are in Sec.~\ref{sec:results}; Fig.~\ref{fig:overview} shows the pipeline.

\begin{figure}[t]
  \centering
  \begin{subfigure}[t]{0.45\linewidth}
    \centering
    \includegraphics[height=2.8cm]{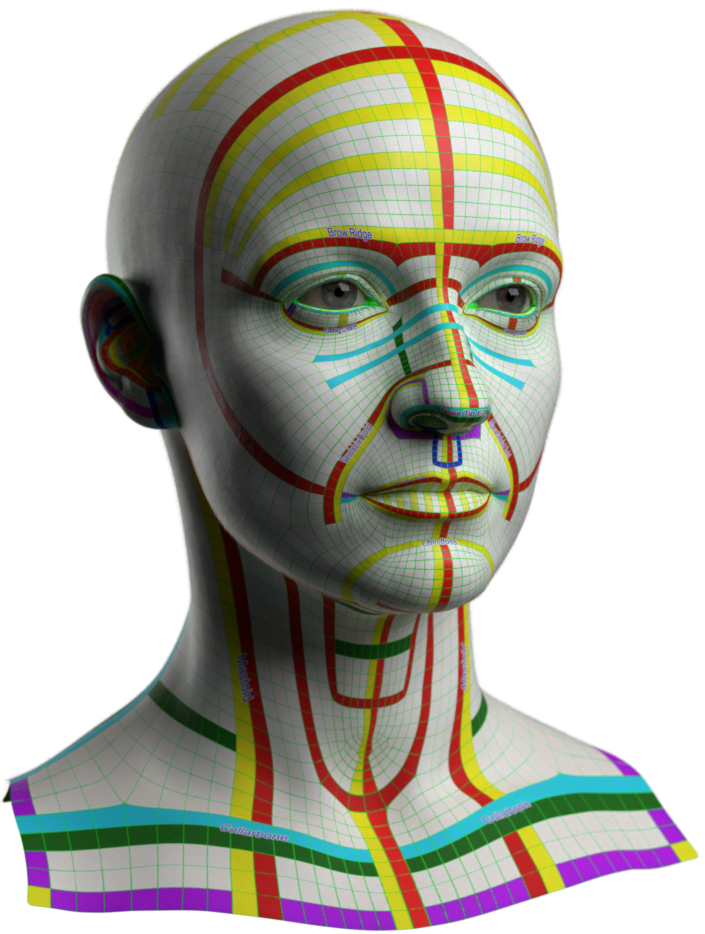}
    \Description{3D rendering of the guide texture.}
    \caption{3D Rendering}
    \label{fig:geo_3d}
  \end{subfigure}\hfill
  \begin{subfigure}[t]{0.45\linewidth}
    \centering
    \includegraphics[height=2.8cm]{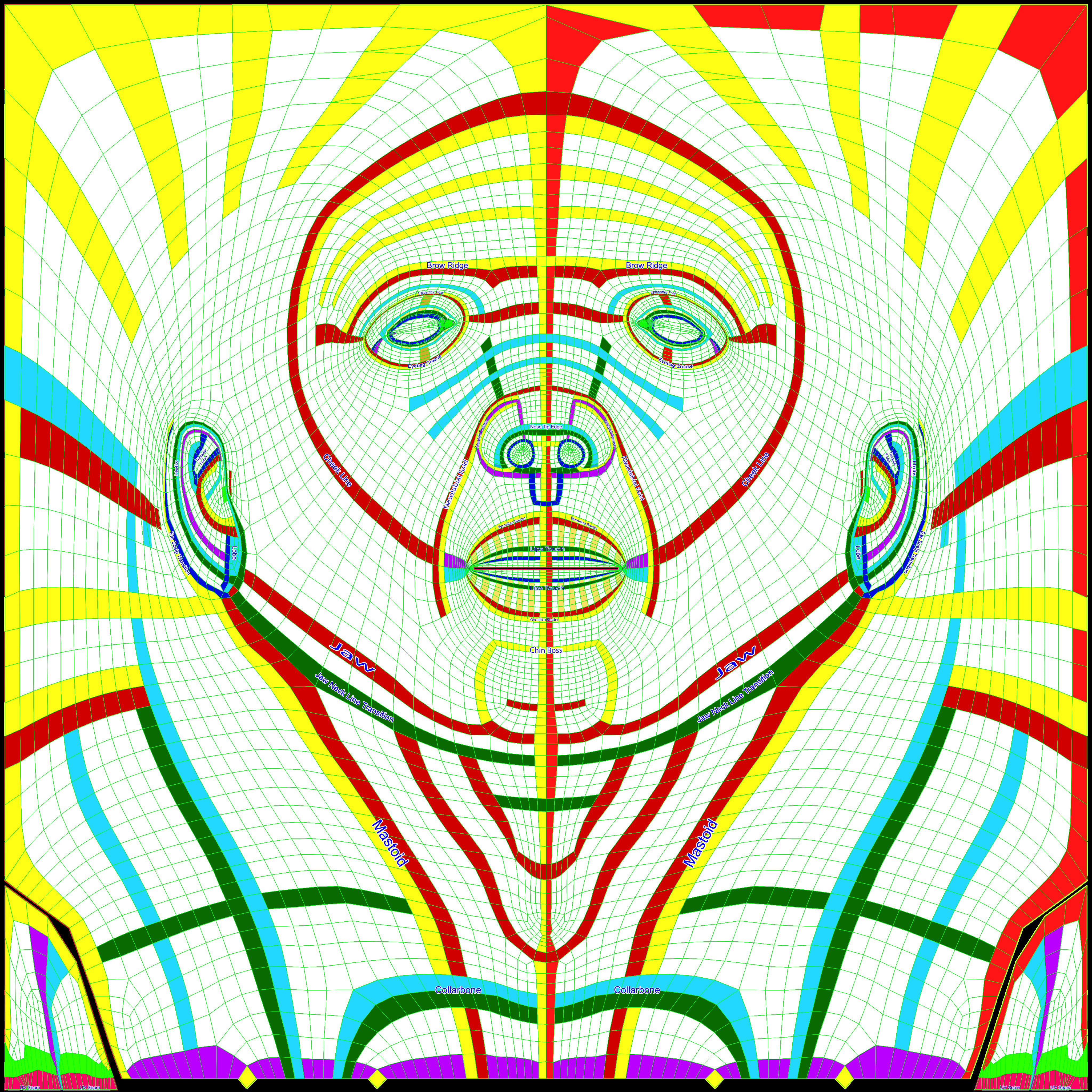}
    \caption{UV Layout}
    \label{fig:geo_uv}
  \end{subfigure}
  \vspace{-1em}
  \caption{Guide texture used for validating mesh alignment.}
  \label{fig:geo_comparison}
\end{figure}

\subsection{UV-Space Data Representation}
\label{method/data}

Multi-view facial data is acquired via polarized spherical gradient illumination~\cite{debevec2000reflectance,ma2007polarized,ghosh2011multiview} and reconstructed with multi-view photogrammetry~\cite{moulon2016openmvg}. We fit a standardized template mesh ($33{,}725$~vertices, $67{,}276$~faces) with Faceform Wrap~\cite{FaceformWrap} and artist-validated alignment using a guide texture (Fig.~\ref{fig:geo_comparison}), yielding a shared UV layout for all modalities.

\textbf{Position maps} are obtained by rasterizing UV triangles with nvdiffrast~\cite{Laine2020diffrast} under conservative rasterization and interpolating per-vertex world-space positions at each covered texel; the result is stored as offsets from a dataset-mean position map (Sec.~\ref{method/mesh}). \textbf{Displacement maps} are computed via Catmull--Clark subdivision~\cite{CatmullClark1978} and per-view normal-image matching~\cite{ghosh2011multiview,graham2013microgeometry}, yielding mesoscopic geometry ($<\!1$~mm); the scalar field is clipped, rescaled, and replicated across RGB channels. \textbf{Albedo} is extracted from cross-polarized images and corrected for self-shadowing and subsurface scattering using Blender/Cycles~\cite{Blender5} correction maps. \textbf{UV normal maps} are rasterized from fitted or predicted meshes and used as ControlNet conditioning signals.

\subsection{VLM-Based Semantic Annotation}
\label{method/captions}

Public 3D facial datasets often lack dense metadata, so we generate structured captions with Gemini~2.5~Pro~\cite{comanici2025gemini}\footnote{Captioning script, prompt template, and JSON schema are provided in the supplement.} from each subject's frontal photograph. A fixed template covers demographics, FACS-based expression, skin micro-detail, and structural geometry; an example caption is:
\begin{quote}\small\itshape
``A 50+ male caucasian with white skin and average build. The expression is surprised. Facial action: Inner and outer brow raisers (AU 1+2) active, creating horizontal forehead furrows; Upper lid raisers (AU 5) active, widening the eyes; \ldots Mature skin with notable forehead wrinkles, fine lines around the eyes, and slight laxity in the jowl and neck area. Oval face shape, elongated skull with a high sloping forehead, moderately defined cheekbones tapering to a rounded jawline.''
\end{quote}
The shared clause structure aligns supervision across tasks. Because captions frequently exceed SDXL's 77-token window, we tokenize without truncation, encode each token stream in 77-token chunks through both CLIP text encoders, and concatenate the per-chunk hidden states along the sequence axis to form the U-Net's text conditioning.

\subsection{Backbone and Geometric Conditioning}
\label{method/backbone}

We adapt SDXL with LoRA~\cite{hu2022lora} on U-Net attention projections ($W_Q, W_K, W_V, W_{\text{out}}$), transformer input/output projections, and ResNet spatial convolutions. Timestep / class / auxiliary embeddings and $\text{conv}_{\text{out}}$ stay frozen; for mesh and displacement we additionally adapt $\text{conv}_{\text{in}}$ to absorb non-image latent statistics. The Normal ControlNet stays frozen for all tasks, reducing overfitting on our $\sim\!2{,}000$~scans. The conditioning scale is fixed at $1.0$ at both training and inference (Sec.~\ref{sec:results_mesh} ablates scale).

At inference, albedo and displacement normals are rendered from FaceRiff's predicted mesh and are smoother than scan normals. To reduce this train/inference shift, we apply stochastic normal smoothing during training: with probability $0.3$, normals are Gaussian-blurred with $\sigma \in [0.5, 2.5]$ and re-normalized to unit length.

\subsection{Residual UV Position-Map Diffusion for Mesh Synthesis}
\label{method/mesh}

Mesh geometry is synthesized as residual diffusion around the dataset-mean UV position map $\bar{P}\in\mathbb{R}^{1024\times1024\times3}$, with per-channel scale $s\in\mathbb{R}^3$ taken as the 99th percentile of $|P-\bar{P}|$ over training texels:
\begin{equation}
\tilde{P} = \mathrm{clamp}\!\left( \tfrac{1}{2}\!\left( \tfrac{P - \bar{P}}{s} + 1 \right),\, 0,\, 1 \right).
\label{eq:offset}
\end{equation}
Per-channel scaling handles facial anisotropy, and the percentile clip limits seam and boundary outliers. At inference, we invert via $\hat{P}=\bar{P}+s\odot(2\hat{\tilde{P}}-1)$ and bilinearly sample the predicted UV position map at each template vertex's UV coordinate, yielding a fixed-topology, identity-specific mesh.

\subsection{Normal-Guided Albedo and Displacement Synthesis}
\label{method/tex}

Albedo and displacement are generated by two LoRA adapters on the shared SDXL backbone, each conditioned on the VLM caption and per-instance UV normals $N$ through the frozen Normal ControlNet. The shared UV layout aligns these with the mesh stage, enabling cascaded inference: during training, $N$ is rasterized from the scan mesh; at inference, from FaceRiff's predicted mesh.
\section{Results}
\label{sec:results}

We evaluate FaceRiff on text-to-mesh synthesis, normal-guided albedo and displacement generation, and end-to-end text-to-3D face generation. We compare with FaceG2E~\cite{wu2023faceg2e} for mesh synthesis and DreamFace~\cite{Zhang2023DreamFace} for rendered appearance, and conduct a blind perceptual study with 45 AAA-studio professionals. Unless otherwise stated, albedo and displacement experiments use ground-truth meshes and normals to isolate texture and mesoscopic-detail quality from mesh-prediction error; Sec.~\ref{sec:results_e2e} evaluates the full cascaded pipeline from text alone.

\subsection{Experimental Setup}
\label{sec:setup}

\paragraph{Dataset.}
Our dataset contains $2{,}049$~facial scans from approximately $350$~identities captured under multiple expressions ($1{,}949$~train / $100$~validation, no identity overlap), sourced from raw scans at $4096~\times~4096$ resolution. Each sample includes 8-bit sRGB albedo, 32-bit displacement in cm, a $1024^2$~UV position map fitted to a shared template mesh ($33{,}725$~vertices, $67{,}276$~faces), the corresponding UV normal map, and a VLM-generated caption. All model training targets and outputs are $1024^2$. Due to consent and licensing constraints, the dataset cannot be publicly released. Dataset statistics $\bar{P}$, $s$, and $\bar{N}$ (Sec.~\ref{method/mesh}) are precomputed from the training set.

\paragraph{Task configurations.}
All tasks use SDXL-base~\cite{podell2024sdxl} with frozen VAE, text encoders, and Normal ControlNet. Mesh uses the dataset-mean UV normal $\bar{N}$; albedo and displacement use per-instance UV normals $N$.

\paragraph{Training.}
Each task trains on a single NVIDIA RTX~A6000 (48\,GB) with AdamW, fp16, batch size~$8$, Min-SNR L1 loss~\cite{hang2023efficient} ($\gamma{=}5$), zero-terminal-SNR~\cite{lin2024common}, and caption dropout $p{=}0.1$ for CFG~\cite{ho2022classifier}. Learning rate is $5{\times}10^{-4}$ with $3{,}000$-step warmup and cosine decay to $10^{-6}$. Mesh, displacement, and albedo train for $10$k, $5$k, and $7$k steps. LoRA adapters use rank $64$, $\alpha=r$, dropout $0.1$, updating only the U-Net ($\sim\!2.93\%$ of parameters).

\paragraph{Inference.}
Inference uses DPM-Solver++ 2M~\cite{lu2022dpm} for $50$ steps at CFG scale $2.0$ (higher values degrade quality) and Normal ControlNet scale $1.0$ (Sec.~\ref{sec:results_mesh}). A batch of $8$ takes $\sim\!2$\,min\,$30$\,s on one A6000, excluding VAE decode and asset export.

\subsection{Text-to-Mesh Synthesis}
\label{sec:results_mesh}

\begin{figure}[t]
  \centering
  \setlength{\tabcolsep}{0pt}
  \renewcommand{\arraystretch}{1.0}
  \footnotesize

  \includegraphics[width=\columnwidth]{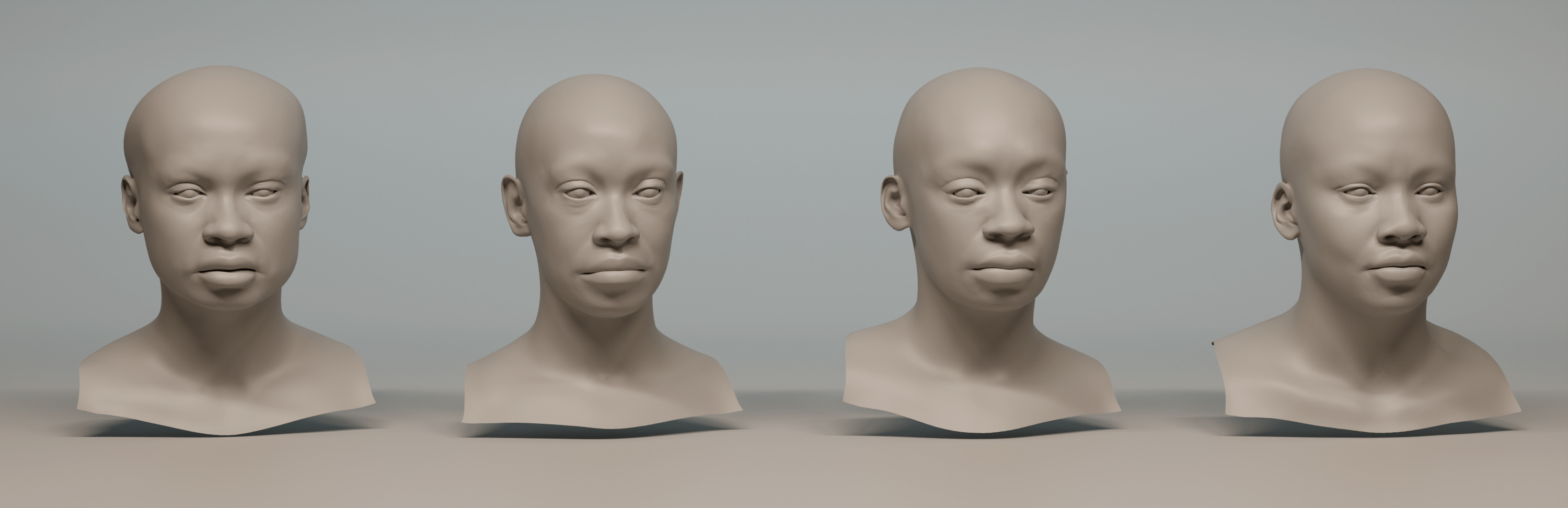}
  \vspace{-8pt}\\
  \textbf{(a)} \itshape A $<$ 24 African \textbf{female} with dark brown skin, average build. Neutral expression: smooth forehead, relaxed eyelids and cheeks, closed lips, frontal gaze. Smooth, evenly pigmented complexion. Oval face with moderate forehead, high cheekbones, softly tapered jawline, and gently projected chin.
  \vspace{6pt}

  \includegraphics[width=\columnwidth]{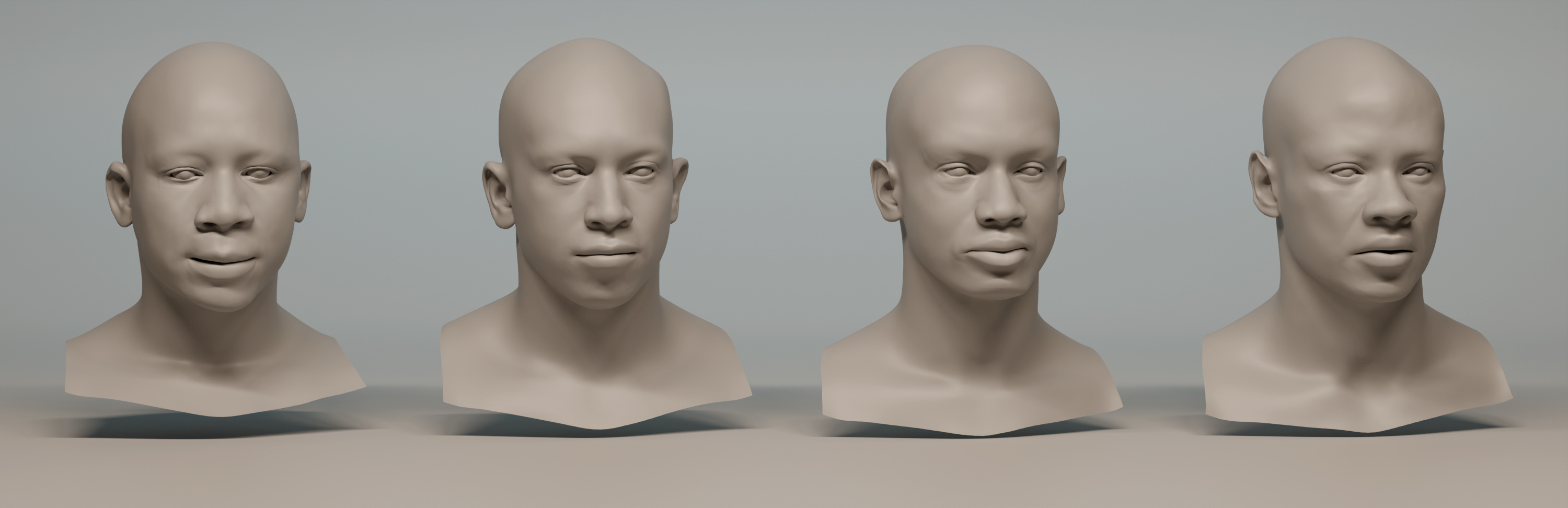}
  \vspace{-8pt}\\
  \textbf{(b)} \itshape Identical to (a), gender clause changed: \emph{female}$\to$\emph{\textbf{male}}.
  \vspace{2pt}

  \includegraphics[width=\columnwidth]{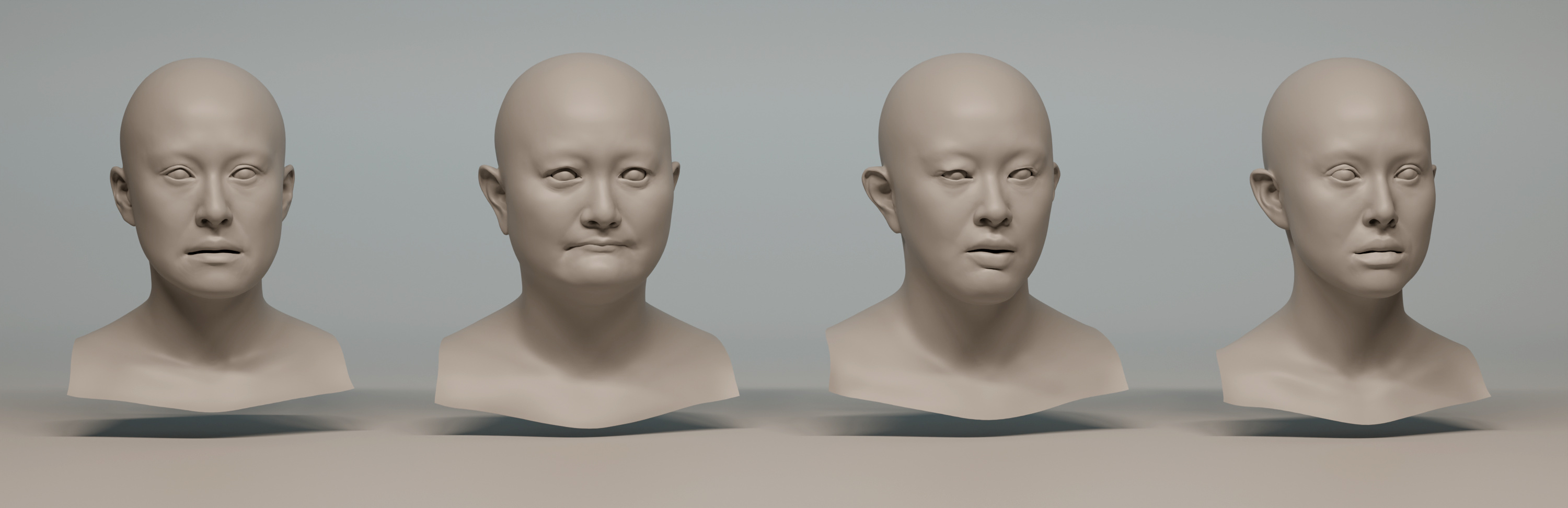}
  \vspace{-8pt}\\
  \textbf{(c)} \itshape A young Asian female.
  \vspace{2pt}

  \includegraphics[width=\columnwidth]{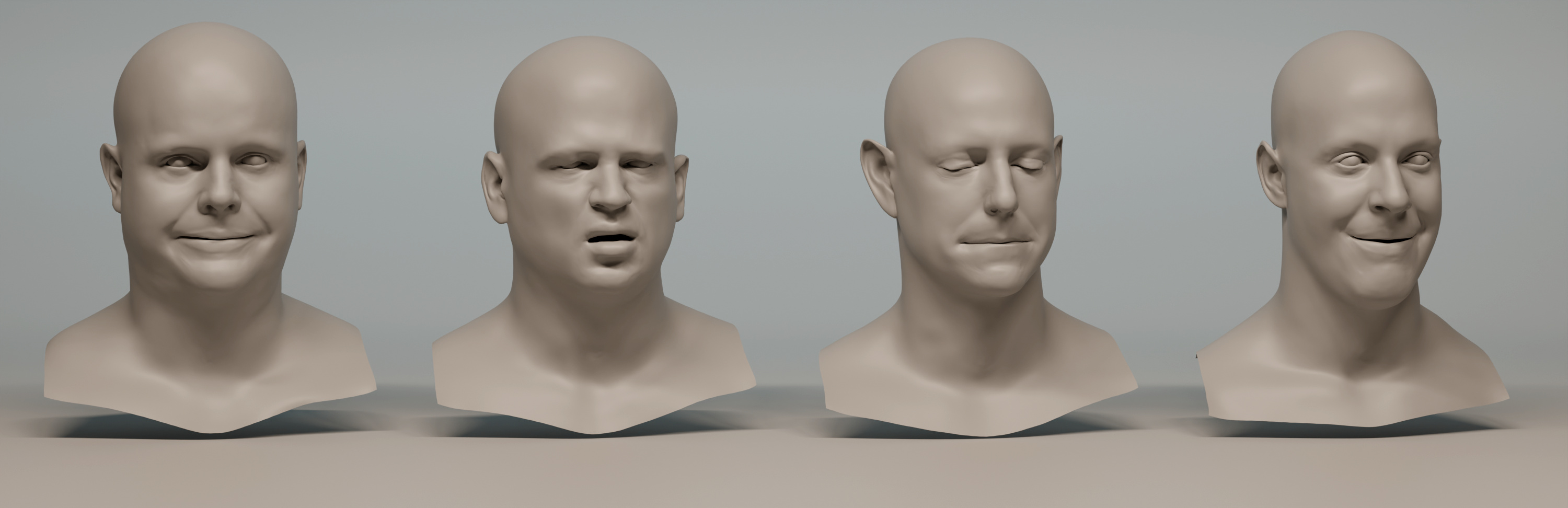}
  \vspace{-8pt}\\
  \textbf{(d)} \itshape A young Caucasian male.
\vspace{-4pt}
\caption{\textbf{FaceRiff text-to-mesh prompt control.}
Four seeds per prompt. Rows (a,b) differ by a single-word gender edit; rows (c,d) use demographic prompts. Identity varies across seeds yet samples stay prompt-faithful; expression is prompted jointly with identity, not edited independently.}

  \Description{FaceRiff mesh generation results for four prompts. Each row shows multiple generated mesh samples from the same text prompt, demonstrating prompt control, gender editing, and seed-level diversity.}
  \label{fig:mesh_prompts}
\end{figure}

Fig.~\ref{fig:mesh_prompts} shows prompt control and seed-level diversity. Rows (a,b) differ only by a gender edit (\emph{female}$\to$\emph{male}); rows (c,d) show diversity from short prompts.

\begin{figure}[t]
  \centering
  \includegraphics[width=\columnwidth]{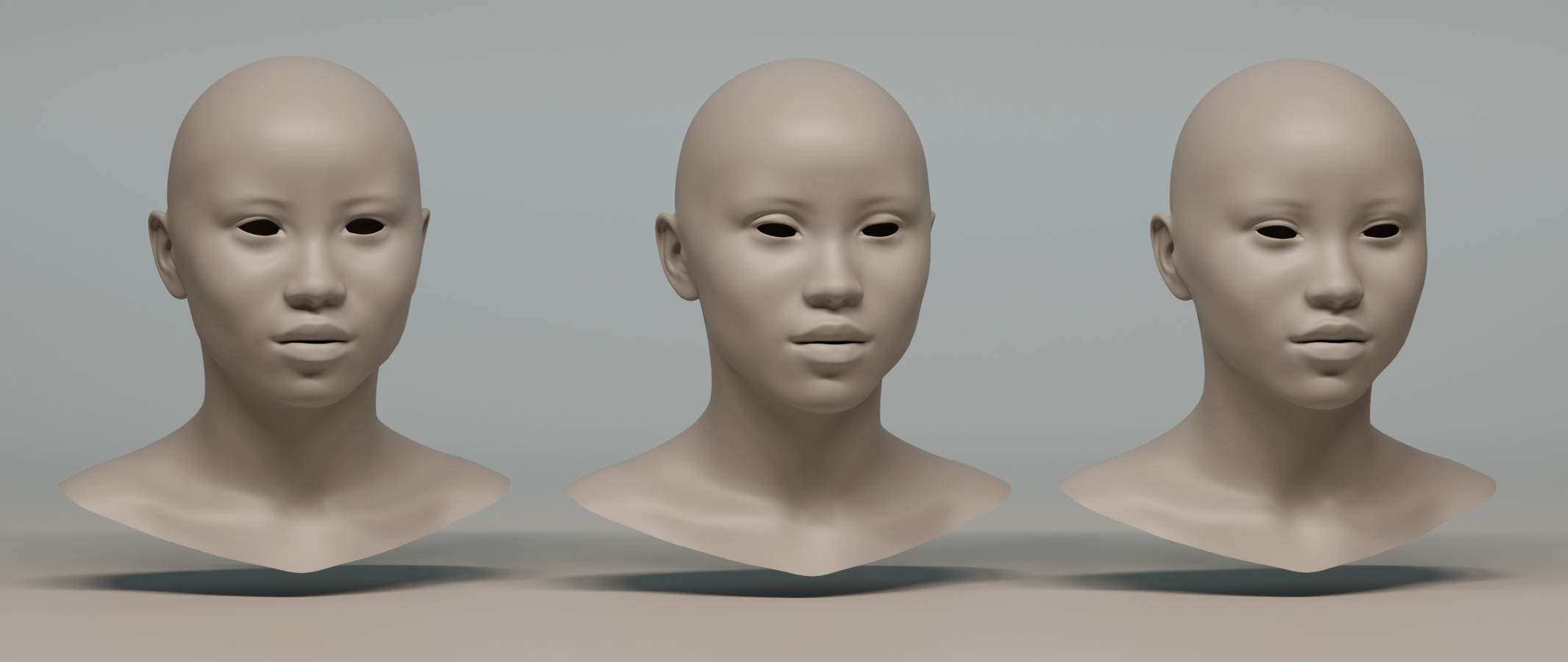}\\
  \includegraphics[width=\columnwidth]{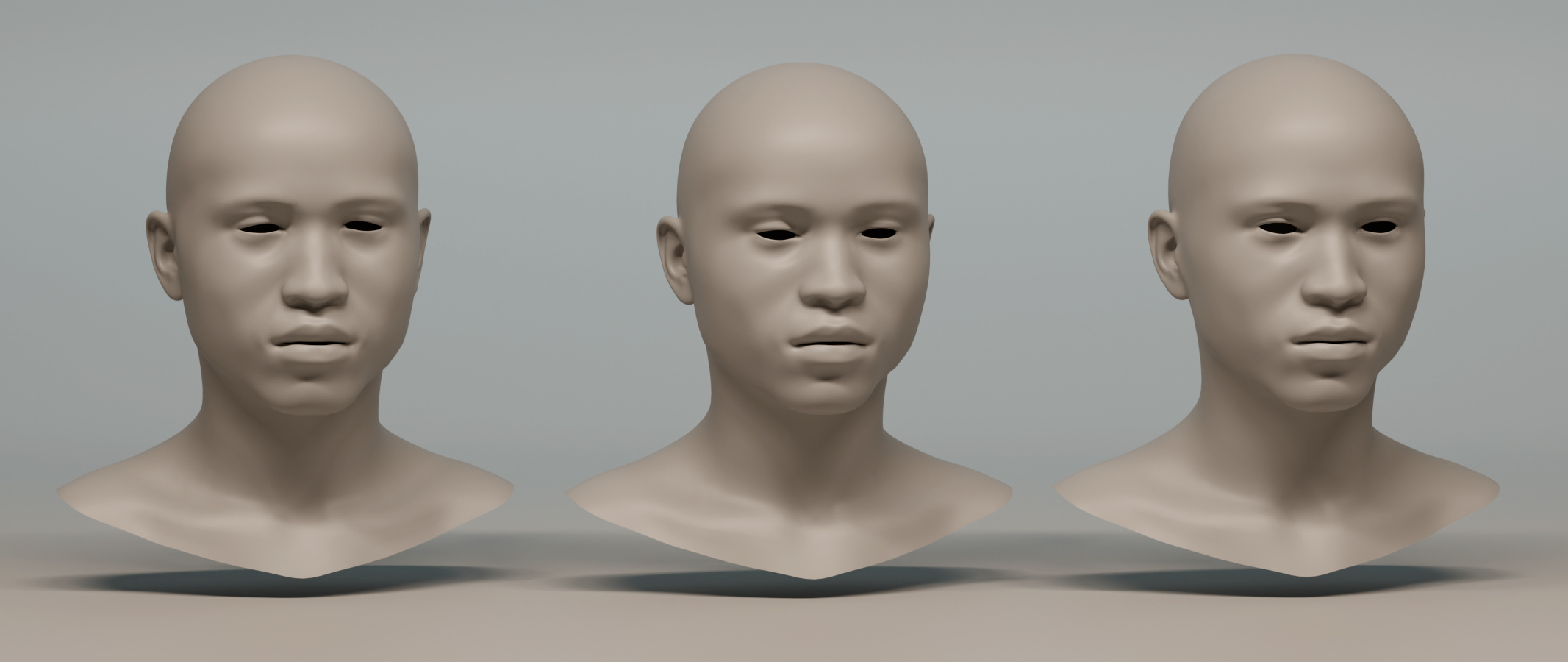}\\
\caption{\textbf{FaceG2E text-to-mesh results.}
Using the first two prompts from Fig.~\ref{fig:mesh_prompts}, we show three random seeds per prompt. FaceG2E produces near-identical meshes across seeds, indicating limited seed-level identity diversity under the same prompt.}
\Description{FaceG2E mesh generation results for the first two prompts from Fig. 4. Each row shows three random seeds, illustrating limited variation across seeds.}
  \label{fig:faceg2e_compare}
\end{figure}

\paragraph{Comparison with FaceG2E}
We compare with FaceG2E~\cite{wu2023faceg2e} using the same prompts and three random seeds per method. Fig.~\ref{fig:faceg2e_compare} shows results for two prompts from Fig.~\ref{fig:mesh_prompts}. Both methods produce plausible prompt-aligned geometry, but FaceG2E yields near-identical samples across seeds, while FaceRiff produces distinct fixed-topology identities compatible with its albedo and displacement branches. FaceG2E's low seed-level variance is consistent with its published SDS setup: short optimization from the HIFI3D mean shape under high guidance tends to converge to the prompt's conditional mode~\cite{Poole2023DreamFusion}.

\subsubsection{Ablation: ControlNet Conditioning Scale}
\label{sec:mesh_ablations}

We evaluate $100$~held-out prompts over the valid face region ($97.6\%$ of UV texels), reporting per-texel mean and 95th-percentile L2 error after denormalizing with $\bar{P}$ and $s$, plus \textsc{DistToMean} and \textsc{Pixel\_Std} as diversity proxies. Because a text prompt does not define a unique ground-truth identity, position error should be interpreted as a coarse distributional fidelity proxy rather than a deterministic reconstruction metric.

Table~\ref{tab:mesh_scale} sweeps the inference ControlNet scale. Conditioning on $\bar{N}$ improves mean error ($12.72 \to 11.47$\,mm) while \textsc{DistToMean} and \textsc{Pixel\_Std} increase, suggesting that the mean normal acts as a spatial scaffold rather than collapsing samples. We adopt scale $1.0$ to match training. Direct comparison to prior text-to-3D face methods is limited because they rarely report text-only mesh-to-GT error, while UV-geometry methods~\cite{otto2024multimodal} use stronger inputs such as portraits, sketches, landmarks, or 3DMM parameters.

\begin{table}[htbp]
\centering
\caption{\textbf{ControlNet scale sweep} ($100$ prompts, mm). Mean error is lowest at scale 1.0; diversity  rises with conditioning strength. Default in \textbf{bold}.}
\label{tab:mesh_scale}
\small
\setlength{\tabcolsep}{5pt}
\begin{tabular}{lcccc}
\toprule
Scale & Mean $\downarrow$ & p95 $\downarrow$ & DistToMean $\uparrow$ & Pixel\_Std $\uparrow$ \\
\midrule
$0.0$ & $12.72$ & $26.57$ & $7.59$ & $0.068$ \\
$0.5$ & $11.63$ & $\mathbf{24.00}$ & $7.58$ & $0.090$ \\
$\mathbf{1.0}$ & $\mathbf{11.47}$ & $24.93$ & $8.10$ & $0.103$ \\
$2.0$ & $12.19$ & $24.64$ & $\mathbf{9.81}$ & $\mathbf{0.123}$ \\
\bottomrule
\end{tabular}
\end{table}

\subsection{Normal-Guided Albedo and Displacement}
\label{sec:results_texture}

We evaluate albedo and displacement with ground-truth UV normals and meshes, isolating texture and mesoscopic-detail quality from mesh error. Additional ablations for this stage are provided in the supplemental material.

\subsubsection{Text-to-Albedo Generation}
\label{sec:results_text2albedo}

\begin{figure}[t]
    \centering
    \setlength{\tabcolsep}{0pt}
    \renewcommand{\arraystretch}{0.95}

    \includegraphics[width=\columnwidth]{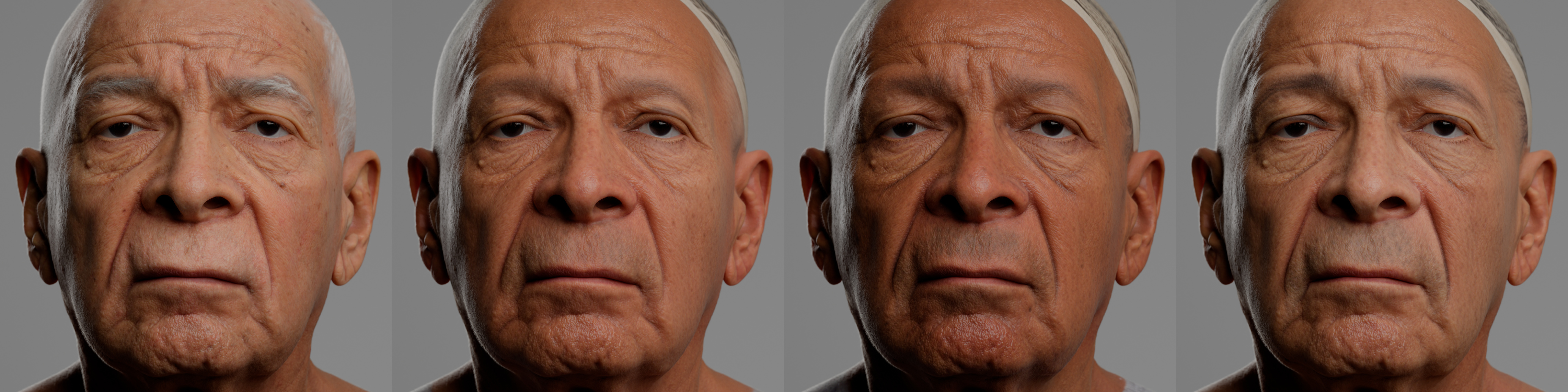}
    \\[0.04cm]
    \includegraphics[width=\columnwidth]{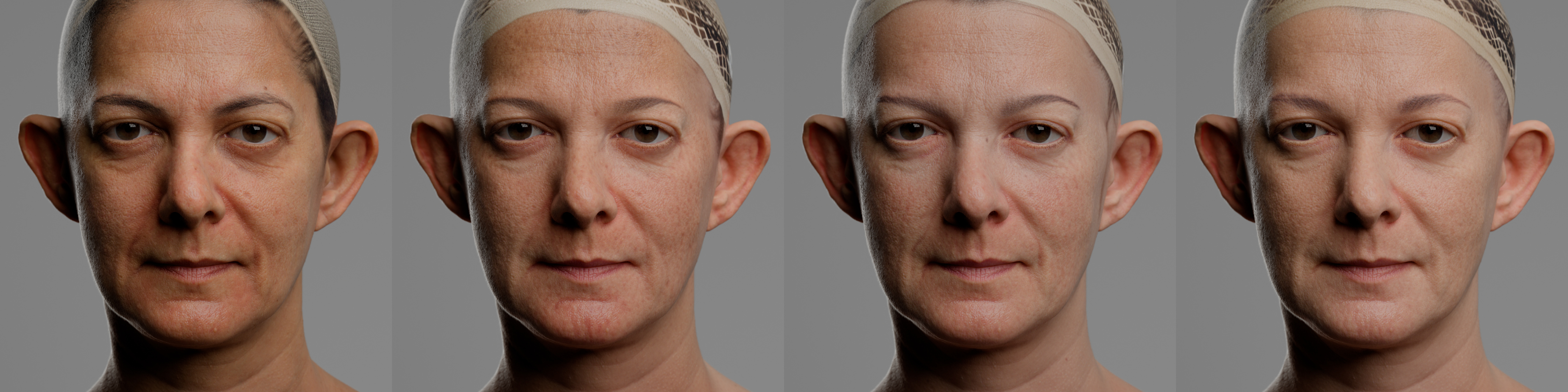}

    \vspace{0.25em}

    {\scriptsize
    \setlength{\tabcolsep}{3pt}
    \begin{tabular}{p{0.10\columnwidth} p{0.84\columnwidth}}
        \textbf{Row 1} &
        \textit{50+ Latino male, light brown skin. Serious neutral expression with slight glabellar furrow (AU 1+2). Mature skin with forehead wrinkles, periorbital fine lines, visible pores, and subtle sun-induced hyperpigmentation. Oval face, rounded jawline.} \\
        \textbf{Row 2} &
        \textit{35-50 Caucasian female, light brown skin. Neutral expression, frontal gaze. Matte skin with fine pores and subtle pigmentation, no prominent wrinkles. Oval face with well-defined cheekbones.}
    \end{tabular}
    }

    \caption{\textbf{Text-to-albedo with GT normal conditioning.} Each row: ground truth and three FaceRiff samples from the same prompt on the same GT mesh. Attributes stay consistent while pore-level detail varies across samples.}
    \Description{Text-to-albedo generation showing two rows of results. Each row contains a ground-truth facial texture and three FaceRiff-generated samples from one prompt, with the corresponding prompts listed underneath.}
    \label{fig:text2albedo_example}
\end{figure}

For each caption, Fig.~\ref{fig:text2albedo_example} shows the ground-truth albedo (first column) alongside three generated samples (columns 2--4) rendered on the GT mesh. Prompted attributes remain consistent across seeds, while pore texture, skin tone, and localized details such as \emph{hyperpigmentation} and \emph{periorbital fine lines} vary plausibly in anatomically appropriate regions.

\paragraph{Ablation: normal conditioning.}
We compare FaceRiff with and without the Normal ControlNet on $100$ validation prompts, keeping other settings fixed. Metrics are computed on UV albedo maps over the valid face region. Table~\ref{tab:albedo_ablation} shows that normal conditioning improves FID, LPIPS, CLIP, and DINO-based distances, supporting its role in localizing text-described pores, wrinkles, and pigmentation.

\begin{table}[htbp]
\centering
\caption{\textbf{Normal conditioning ablation} for text-to-albedo on UV albedo maps ($100$ validation prompts, valid face region). Default in \textbf{bold}.}
\label{tab:albedo_ablation}
\small
\setlength{\tabcolsep}{4pt}
\begin{tabular}{lccccc}
\toprule
Config & FID $\downarrow$ & LPIPS $\downarrow$ & CLIP $\uparrow$ & DINO$_\text{pair}$ $\downarrow$ & DINO$_\text{FD}$ $\downarrow$ \\
\midrule
w/o normal   & $85.01$ & $0.338$ & $0.290$ & $0.252$ & $621.79$ \\
\textbf{w/ normal} & $\mathbf{80.71}$ & $\mathbf{0.316}$ & $\mathbf{0.295}$ & $\mathbf{0.236}$ & $\mathbf{613.64}$ \\
\bottomrule
\end{tabular}
\end{table}

\begin{figure}[htbp]
    \centering
    \setlength{\tabcolsep}{1pt}
    \begin{minipage}{0.3\columnwidth}\centering
        \includegraphics[width=\linewidth]{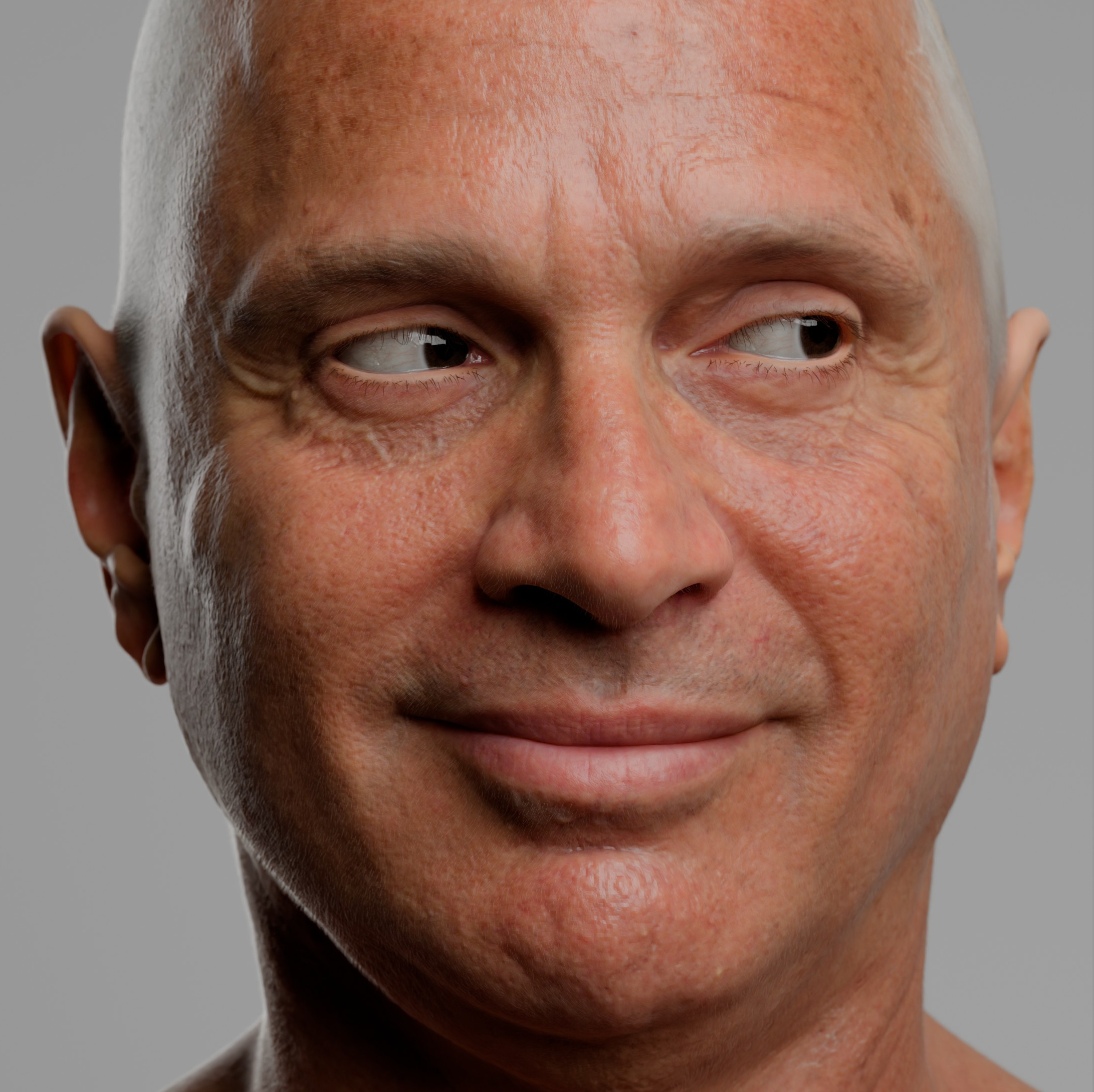}\\[1mm]
        \includegraphics[width=\linewidth]{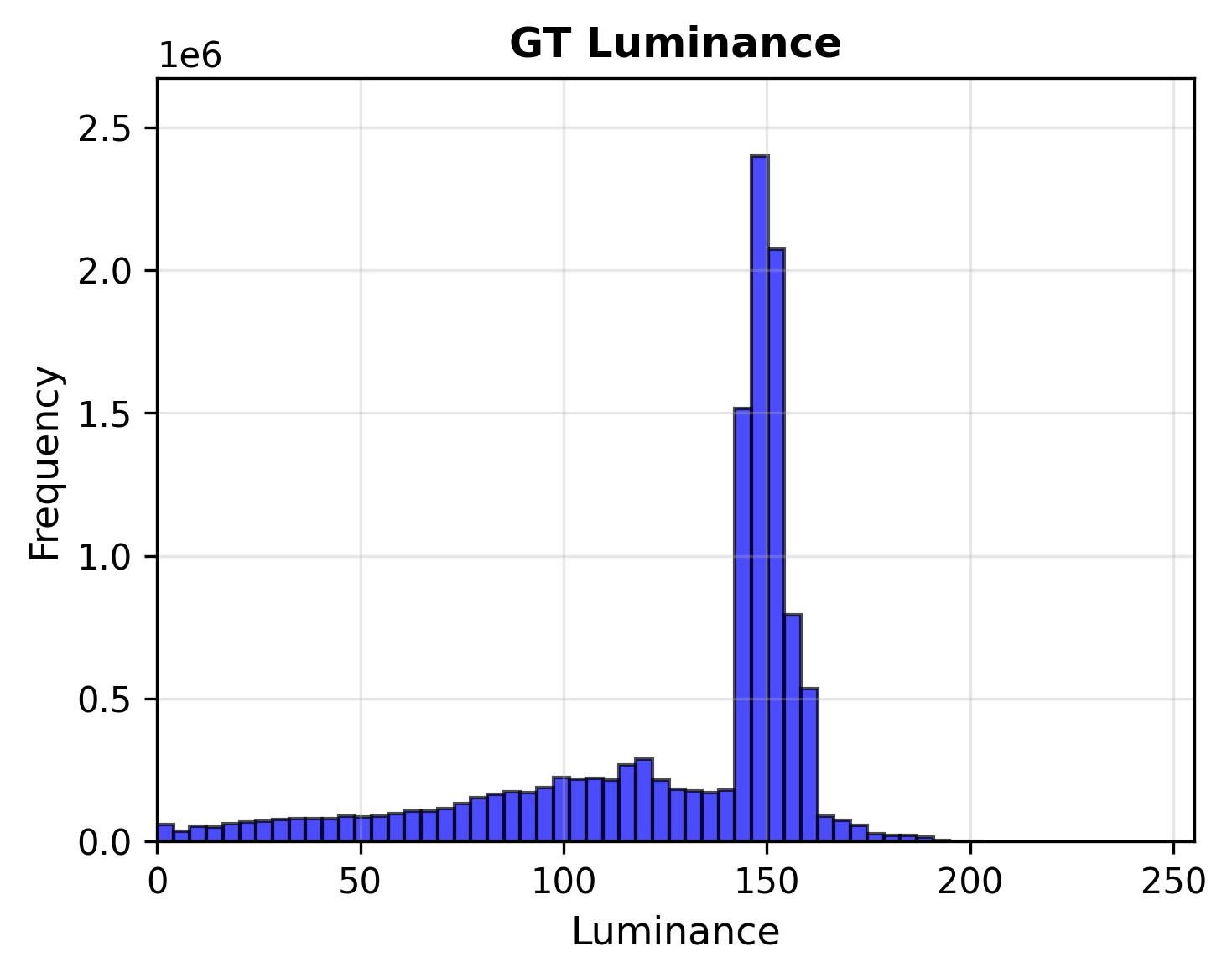}\\
        {\small (a) GT}
    \end{minipage}\hfill
    \begin{minipage}{0.3\columnwidth}\centering
        \includegraphics[width=\linewidth]{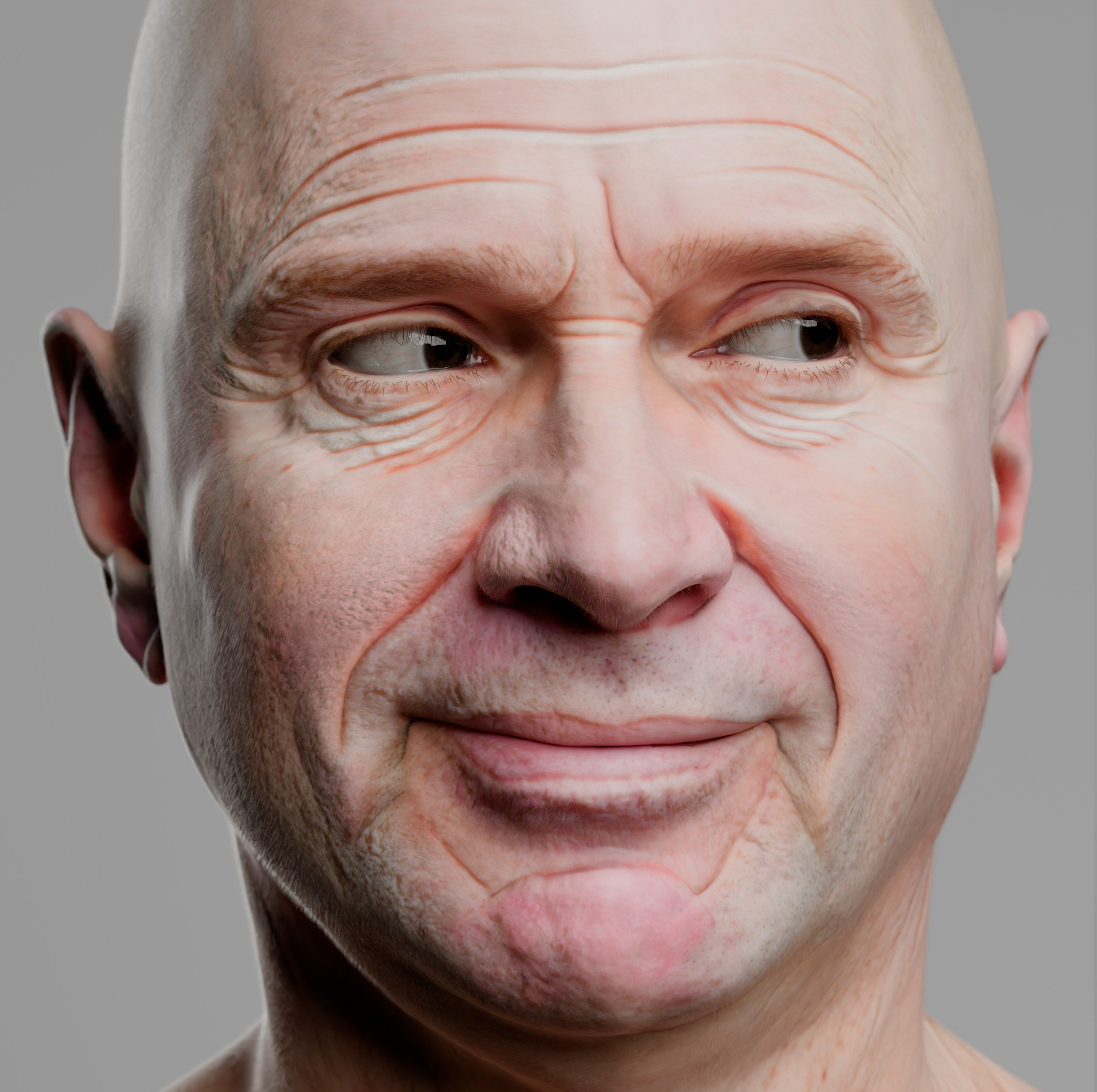}\\[1mm]
        \includegraphics[width=\linewidth]{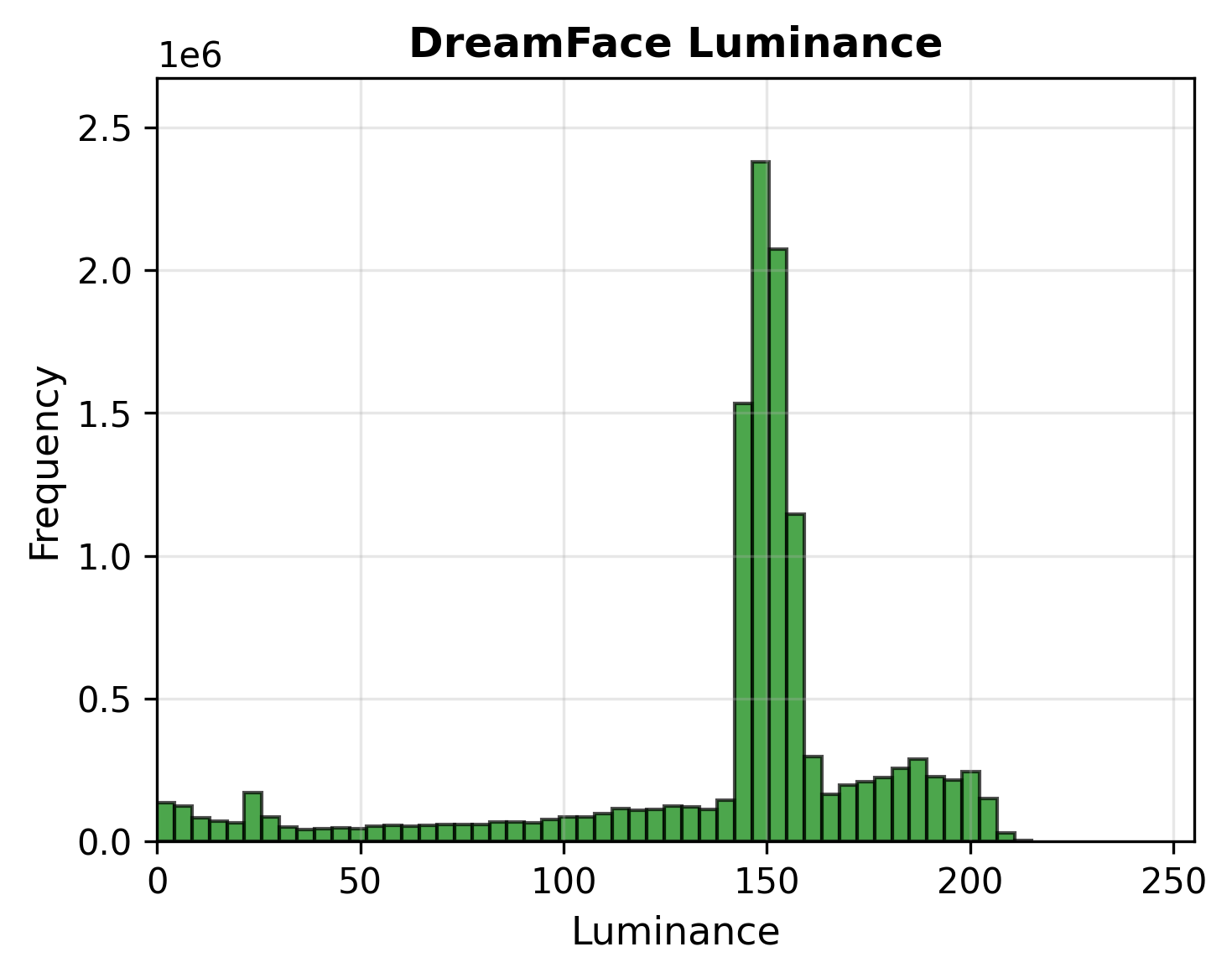}\\
        {\small (b) DreamFace}
    \end{minipage}\hfill
    \begin{minipage}{0.3\columnwidth}\centering
        \includegraphics[width=\linewidth]{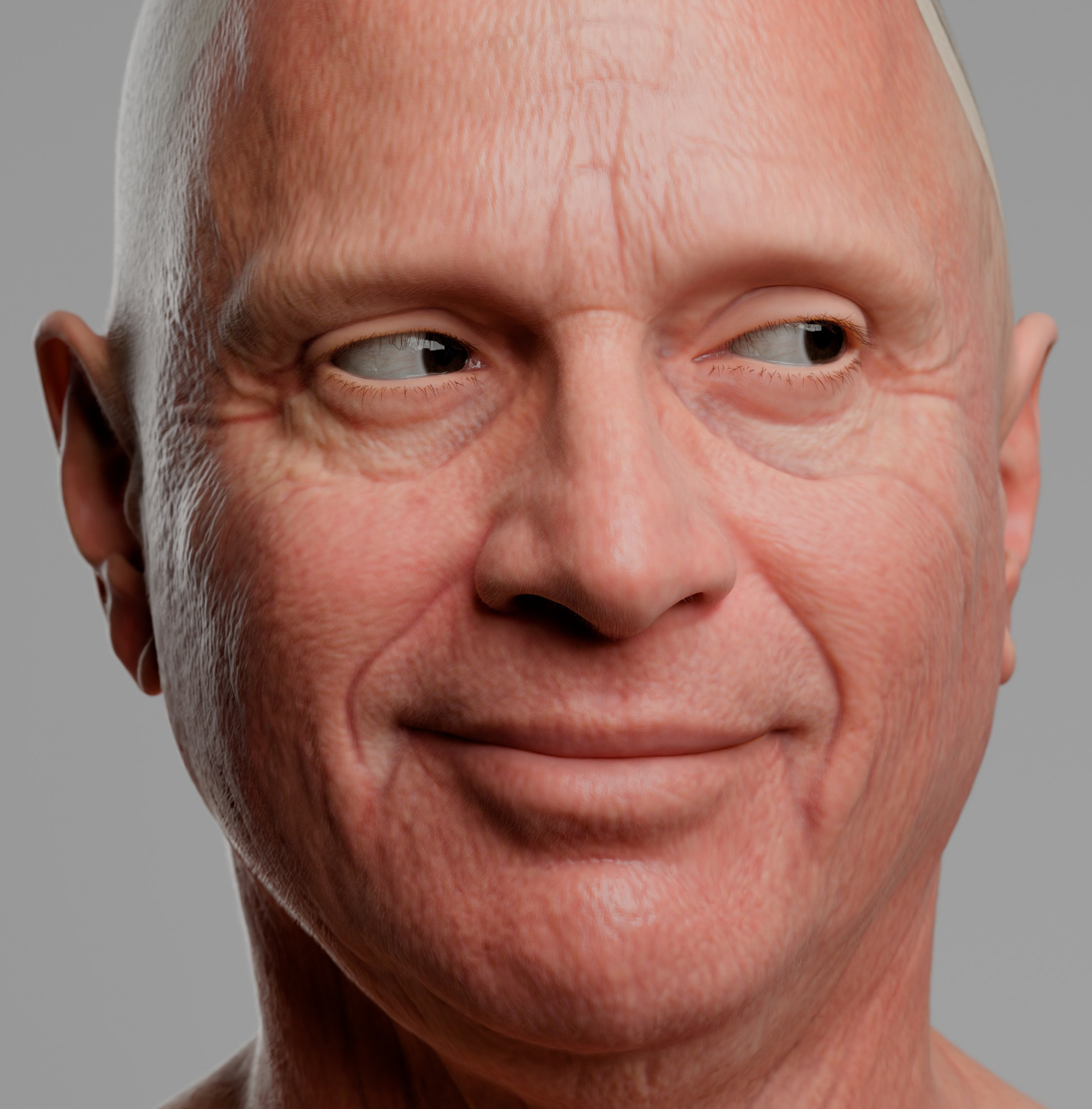}\\[1mm]
        \includegraphics[width=\linewidth]{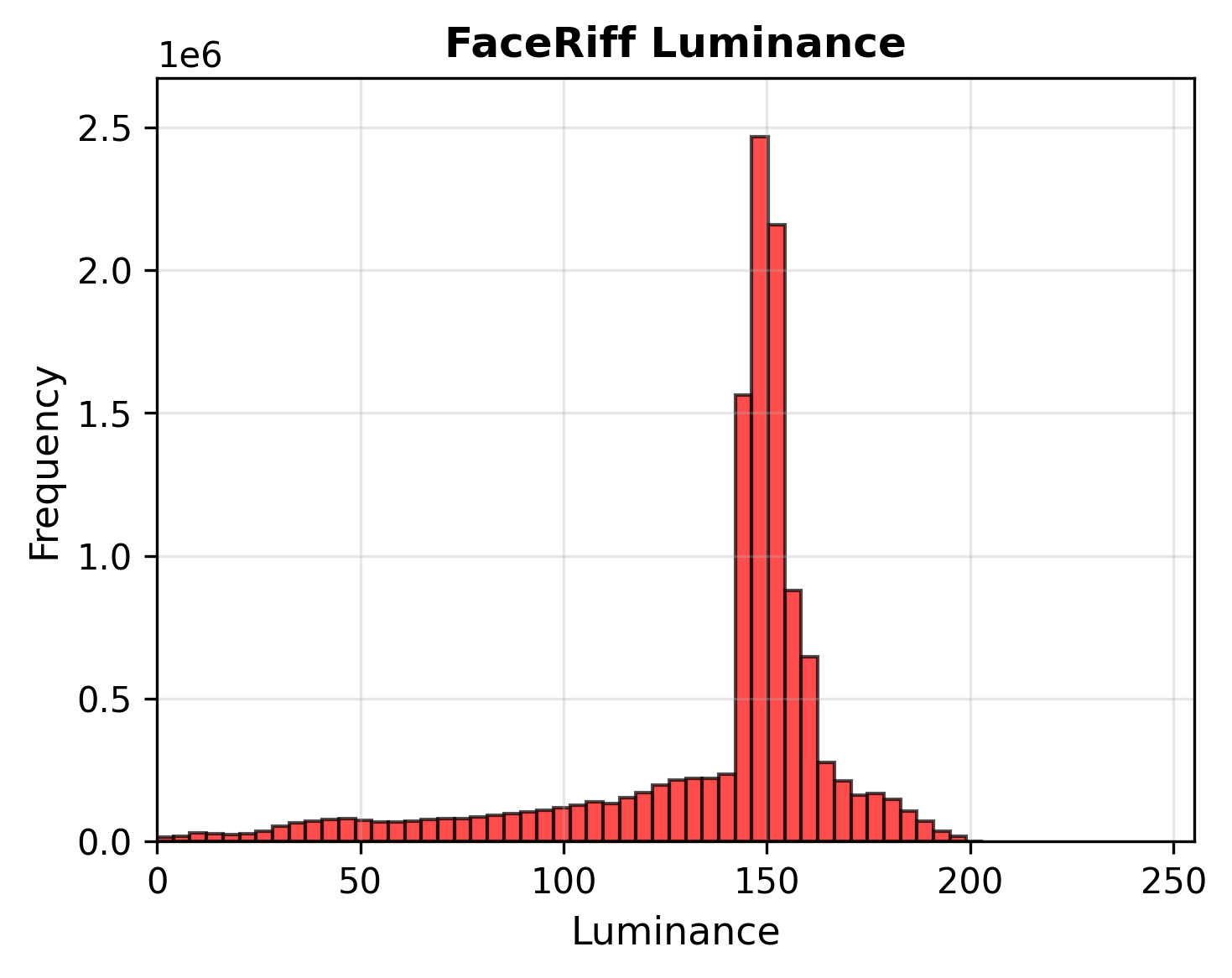}\\
        {\small (c) FaceRiff}
    \end{minipage}

    \caption{\textbf{Luminance distribution comparison.} DreamFace (b) shows an extended high-luminance tail from baked-in lighting; FaceRiff (c) more closely matches the GT distribution (a). See Fig.~\ref{fig:dreamface_comparison} for full qualitative comparisons.}
    \label{fig:luminance_comparison}
\end{figure}

\paragraph{Comparison with DreamFace.}
We compare with DreamFace~\cite{Zhang2023DreamFace} on identical prompts and GT meshes, remapping its public-service outputs to our UV layout via mesh correspondences and UV-space ARAP~\cite{sorkine2007rigid} without altering color, gamma, or tone mapping. As DreamFace lacks UV normals, we use our no-normal variant. Both are rendered under identical Blender/Cycles color management (sRGB, Rec.709, shared tone-mapping/ODT) calibrated rather than tuned to either method, so DreamFace's localized shading and high-luminance artifacts are baked into its textures, not introduced by our pipeline. On a 20-prompt frontal-render set, FaceRiff improves CLIP ($36.3$ vs.\ $35.4$) and FID ($56.4$ vs.\ $107.2$).\footnote{Render-based CLIP scores are not directly comparable to the UV-map CLIP scores in Table~\ref{tab:albedo_ablation}, which use UV albedo maps on a different 100-prompt set.} Consistent with this, Fig.~\ref{fig:luminance_comparison} shows DreamFace's extended high-luminance tail while FaceRiff matches GT, and Fig.~\ref{fig:dreamface_comparison} shows more lighting-independent albedo, sharper pores, and greater diversity.

\subsubsection{Text-to-Displacement Generation}
\label{sec:results_text2disp}

\begin{figure}[!htbp]  
    \centering  
    \begin{tabular}{c}  
      \includegraphics[width=\columnwidth]{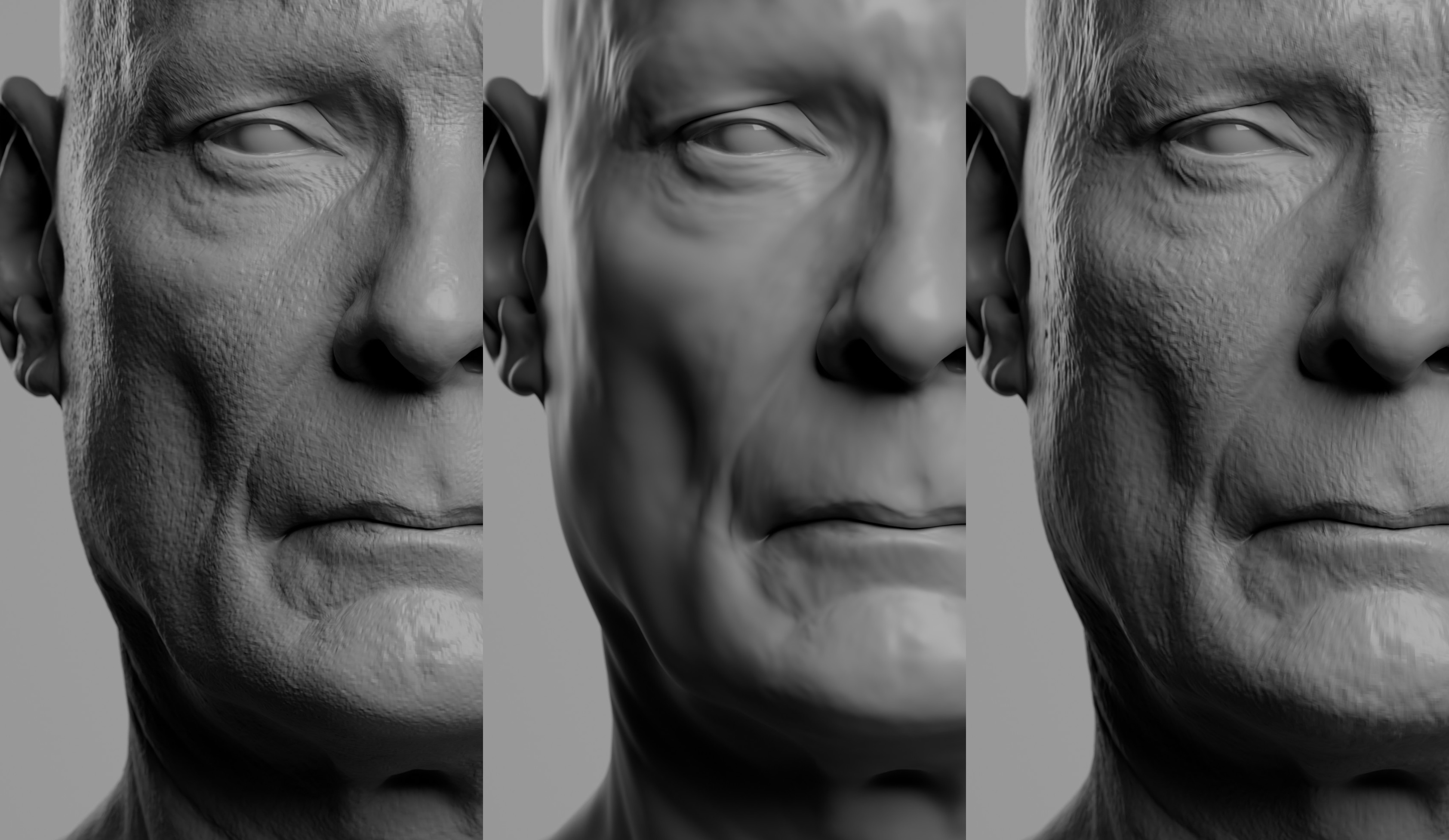} \\
      \includegraphics[width=\columnwidth]{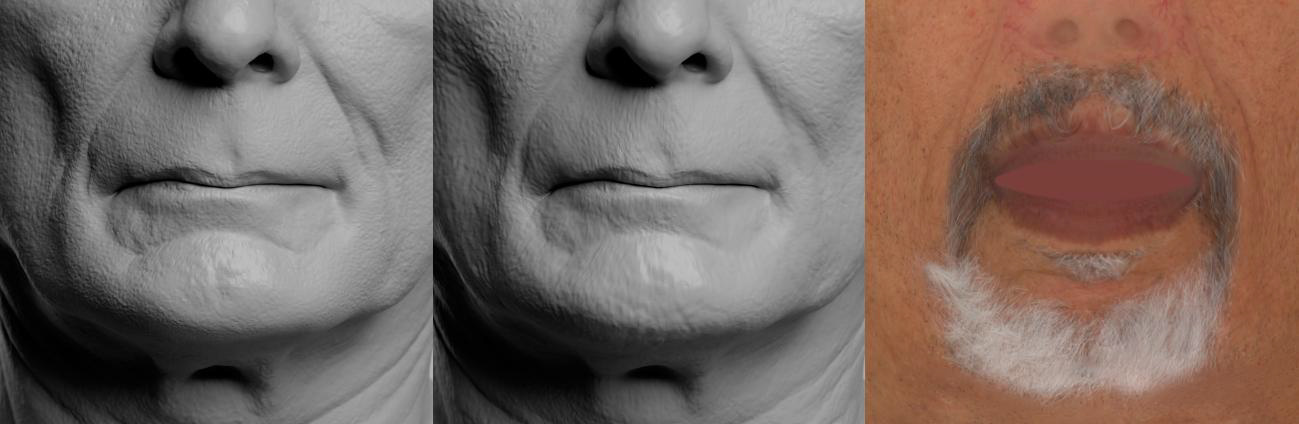}
    \end{tabular}  
    \vspace{-8pt}
    \caption{\textbf{Text-to-displacement with normal conditioning.} Top: GT displacement, base mesh without displacement, and FaceRiff-generated displacement. Bottom: beard-region close-up showing that FaceRiff synthesizes underlying skin detail rather than displacing facial hair.}
    \Description{Two-row displacement result. Top row: ground-truth displacement, mesh without displacement, and FaceRiff-generated displacement. Bottom row: beard-region close-up showing FaceRiff synthesizes skin detail beneath facial hair rather than displacing the hair.}
    \label{fig:disp}  
\end{figure}
\begin{figure*}[!htbp]
    \centering
    \setlength{\tabcolsep}{2pt}
    \renewcommand{\arraystretch}{1.0}

    \begin{tabular}{ccccc}
        \includegraphics[height=0.18\textheight]{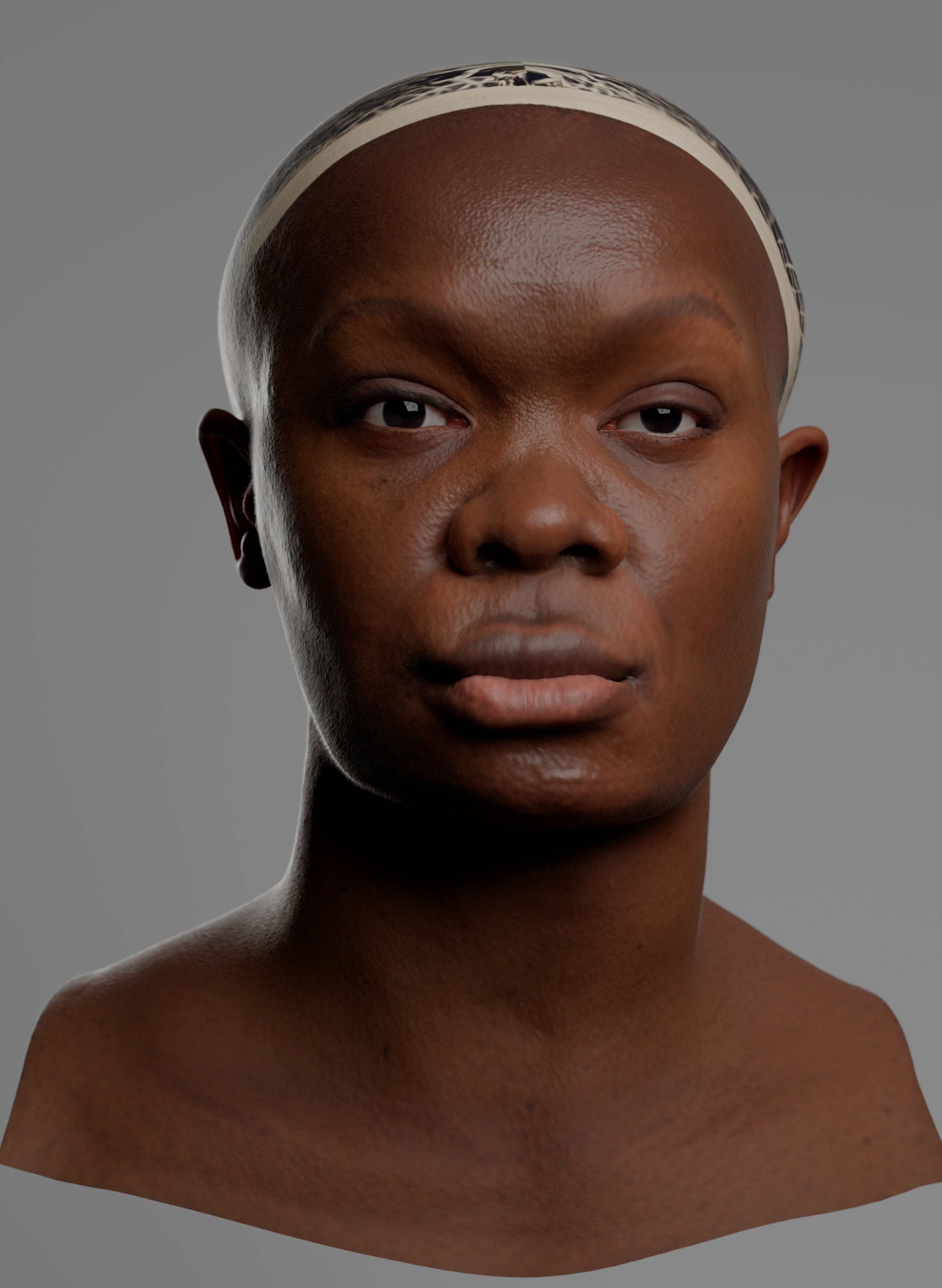} &
        \includegraphics[height=0.18\textheight]{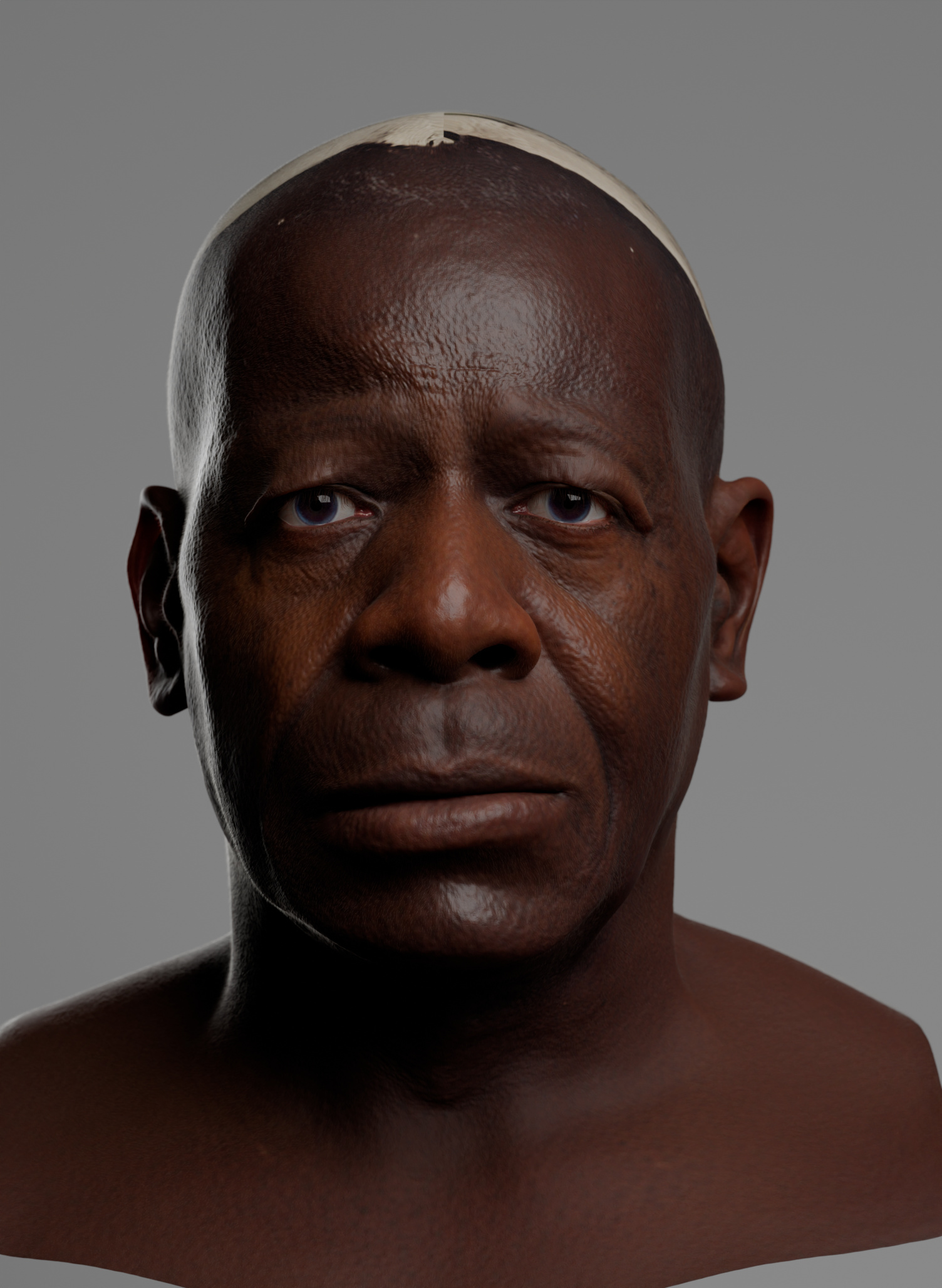} &
        \includegraphics[height=0.18\textheight]{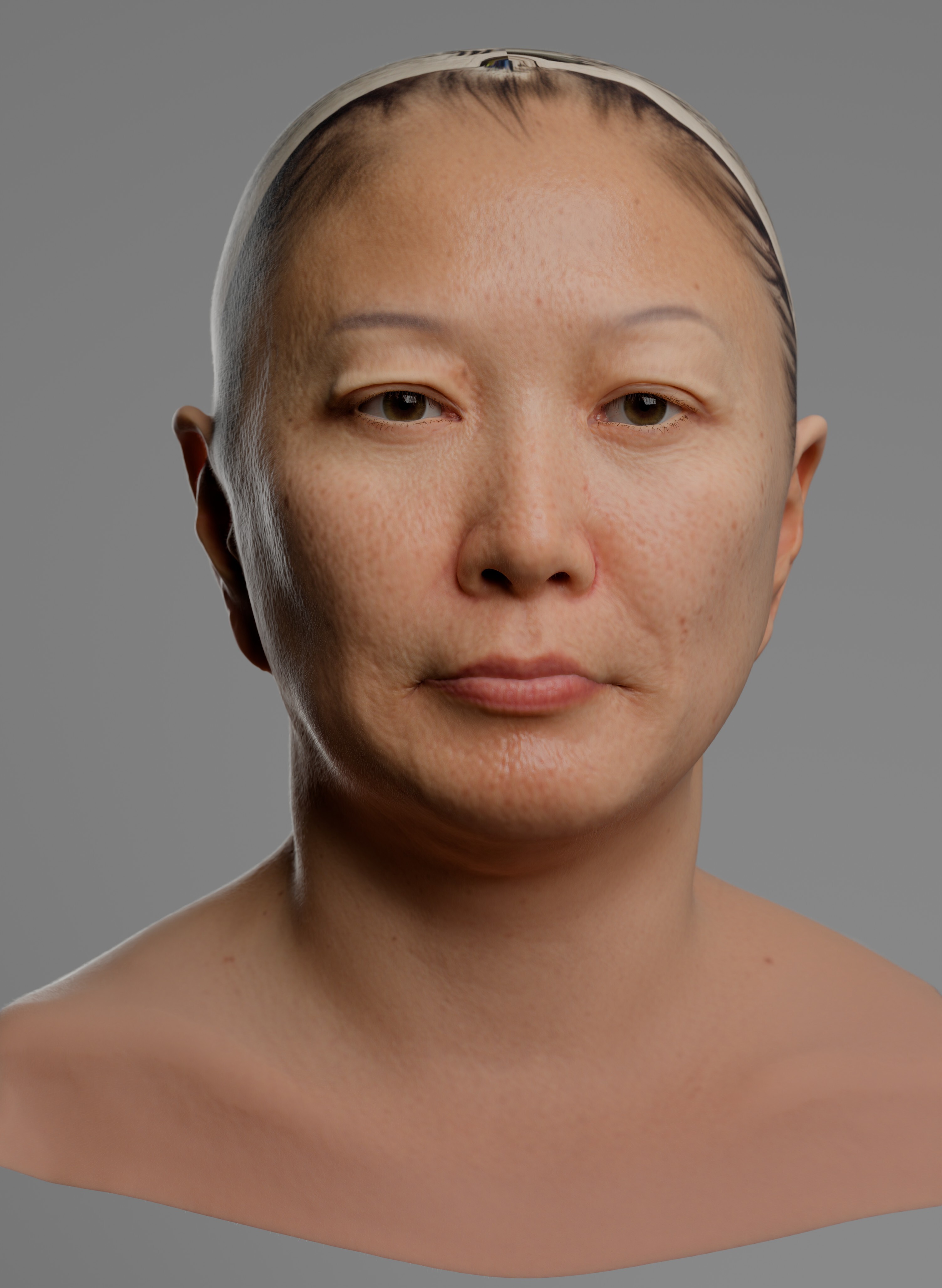} &
        \includegraphics[height=0.18\textheight]{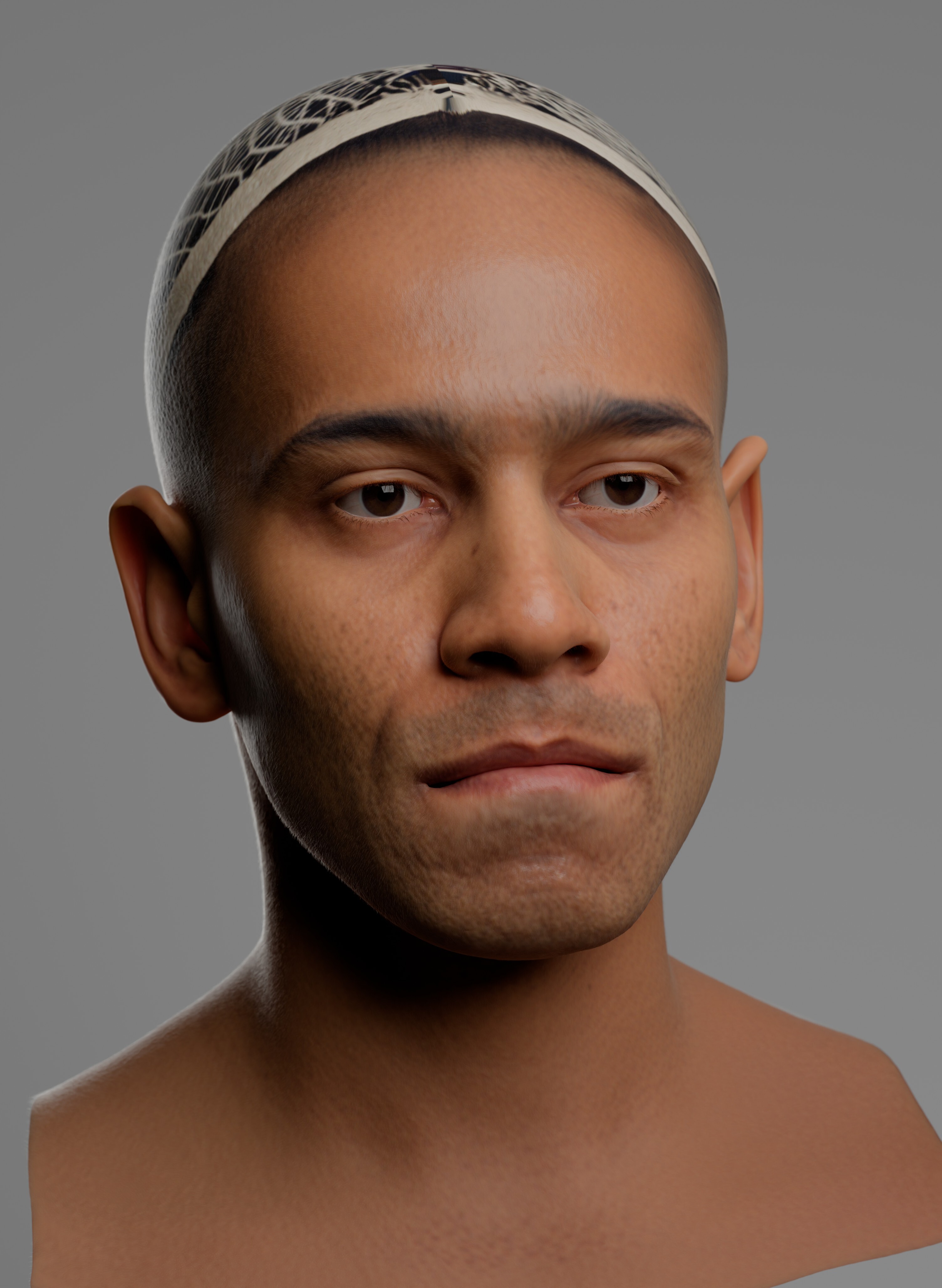} \\
        \scriptsize (a) & \scriptsize (b) & \scriptsize (c) & \scriptsize (d) 
    \end{tabular}

    {\scriptsize
    \setlength{\tabcolsep}{4pt}
    \renewcommand{\arraystretch}{1.12}
    \begin{tabularx}{0.98\textwidth}{p{0.045\textwidth} X}
    \textbf{(a)} &
    \textit{A $<$24 African female, dark brown skin, neutral composed expression, smooth clear complexion, oval face with high cheekbones and softly rounded jawline.} \\[0.15em]

    \textbf{(b)} &
    \textit{A 50+ African male, dark brown skin, slight facial tension with wide eyes and pressed lips, fine periorbital wrinkles, bald scalp, broad forehead, and rounded chin.} \\[0.15em]

    \textbf{(c)} &
    \textit{A 35--50 Asian female, light brown skin, neutral expression with firm closed lips, subtle freckling on nose and cheeks, round face with soft jawline and rounded chin.} \\[0.15em]

    \textbf{(d)} &
    \textit{A 25--34 male, light brown skin, neutral composed expression, smooth skin with visible pores and slight temple discoloration, oval face with subtle cheekbones and short stubble.}
    \end{tabularx}
    }

    \vspace{-0.2em}
    \caption{\textbf{End-to-end 3D face generation.}
    Four full-head renders combining FaceRiff-generated mesh, albedo, and displacement, each produced from text alone. Mesh is synthesized first, and its UV normal map conditions albedo and displacement, preserving cross-modality correspondence without post-hoc registration. Prompt summaries are shown below; full prompts are provided in the supplemental material. Results span diverse demographics and fine-grained attributes (ethnicity, age, skin tone, and skin detail).}
    \label{fig:e2e_results}
\end{figure*}
Fig.~\ref{fig:disp} shows representative displacement results, including full-face comparisons and a facial-hair close-up. FaceRiff captures age-related roughness, expression creases, and memory wrinkles absent from the base mesh. In facial-hair regions, it synthesizes plausible underlying skin detail rather than displacing the hair itself.

Since text prompts do not define a unique ground-truth displacement, we evaluate displacement via rendered appearance and the expert study in Sec.~\ref{sec:results_user_study}; generated details (lip creases, periorbital wrinkles) align with anatomical regions despite UV distortions.

\subsubsection{Expert Perceptual Evaluation}
\label{sec:results_user_study}

We conduct a blind perceptual study with $45$ industry professionals, including Character Directors, Rendering Engineers, and Character Artists. Participants viewed randomized renderings without method labels. Matched GT meshes isolate albedo and displacement quality from mesh-prediction error.

Ground-truth scans were judged real by approximately $90\%$ of participants. FaceRiff outputs were judged real more often than DreamFace outputs: $48.9\%$--$71.1\%$ across examples (mean $59.4\%$) versus $2.2\%$--$17.8\%$ (mean $8.9\%$). On production-oriented criteria using a Poor/Fair/Good/Excellent scale, $70.0\%$ of raters judged FaceRiff's prompt alignment as Good or Excellent, and $61.6\%$ judged its displacement realism as Good or Excellent, with fewer than $6\%$ Poor ratings in either category. These results indicate higher perceived realism than DreamFace in the matched-geometry setting, along with favorable ratings for prompt alignment and displacement realism.

\subsection{End-to-End 3D Face Generation}
\label{sec:results_e2e}

The preceding sections isolate each task with GT conditioning. Here we run the full FaceRiff pipeline from text alone: we first generate the mesh (Sec.~\ref{sec:results_mesh}), compute its per-instance UV normal map $N$, and use $N$ to condition albedo and displacement synthesis. All three modalities are produced from the \emph{same} text prompt. Fig.~\ref{fig:e2e_results} shows full-head renders combining the generated mesh, albedo, and displacement. The shared UV layout provides structural cross-modality correspondence, with no seam correction or optimization step required between tasks; because albedo and displacement use separate stochastic branches, we do not claim exact correspondence for every individual pore, though we observe no visually obvious cross-modal artifacts. Because these branches are conditioned on normals rendered from the generated mesh, mesh errors can propagate to later stages; we mitigate this train/inference shift with the stochastic normal smoothing of Sec.~\ref{method/backbone}, and the shared UV representation preserves cross-modality registration throughout the cascade.

\subsection*{Ethics, data, and bias.}
Our scans were collected with informed consent for machine-learning research. Due to consent and licensing constraints, the dataset cannot be publicly released. VLM captions are reviewed by human evaluators to reduce annotation errors and demographic bias. We acknowledge the dual-use risk of realistic face-asset synthesis; FaceRiff is intended for controlled production workflows, trained only on consented data, and should be deployed with appropriate provenance, licensing, and misuse-prevention safeguards.

\section{Conclusion}
We presented FaceRiff, a data-efficient unified framework that adapts a pretrained SDXL backbone with task-specific LoRA modules to synthesize mesh, albedo, and displacement from text in a shared UV space. Its novelty lies not in SDXL, LoRA, or ControlNet individually, but in organizing them into a unified feed-forward framework with asymmetric geometric conditioning. Three design choices underpin this: residual position offsets from a dataset-mean anchor, per-instance normals for texture modalities with a dataset-mean normal for mesh, and dense VLM captions for semantic control. Together these yield pore-level detail and structural cross-modality correspondence from only $2{,}049$~scans while updating $2.93\%$ of parameters per task. In a blind study with AAA-studio professionals, FaceRiff was judged real substantially more often than DreamFace in matched-geometry renderings, highlighting its practical value and visual quality. FaceRiff shows that geometric conditioning and parameter-efficient adaptation can make pretrained image diffusion models effective for specialized 3D face asset synthesis under limited data, reducing the cost of high-quality face assets.

\bibliographystyle{ACM-Reference-Format}
\bibliography{bibliography}
\newlength{\dreamfaceimgheight}
\setlength{\dreamfaceimgheight}{2.0cm}
\newcommand{\dreamfaceimg}[1]{\includegraphics[height=\dreamfaceimgheight]{#1}}

\begin{figure*}[!ht]
    \centering
    \textbf{Prompt 1:} \\
    \small\textit{``A 35-50 Caucasian male with white skin and an average build. The expression is other/ambiguous, suggesting a state of skeptical contemplation. Facial analysis: Frontalis muscle is active, raising the brows and creating horizontal wrinkles (AU 1+2)., Upper lids are raised (AU 5), with eyes looking upwards., Lips are pressed together with a slight upward curve, forming a subtle smirk., Head is held straight, eyes are directed up and slightly to the right (AU 63).. The skin shows signs of sun damage, including light freckling and uneven pigmentation. Horizontal wrinkles are present on the forehead, and fine lines are visible around the eyes. There are a few small blemishes and visible pores, particularly in the t-zone and chin area.''} \\[0.2cm]
    \begin{tabular}{@{}c@{\hspace{0.01\textwidth}}c@{\hspace{0.01\textwidth}}c@{}}
        \dreamfaceimg{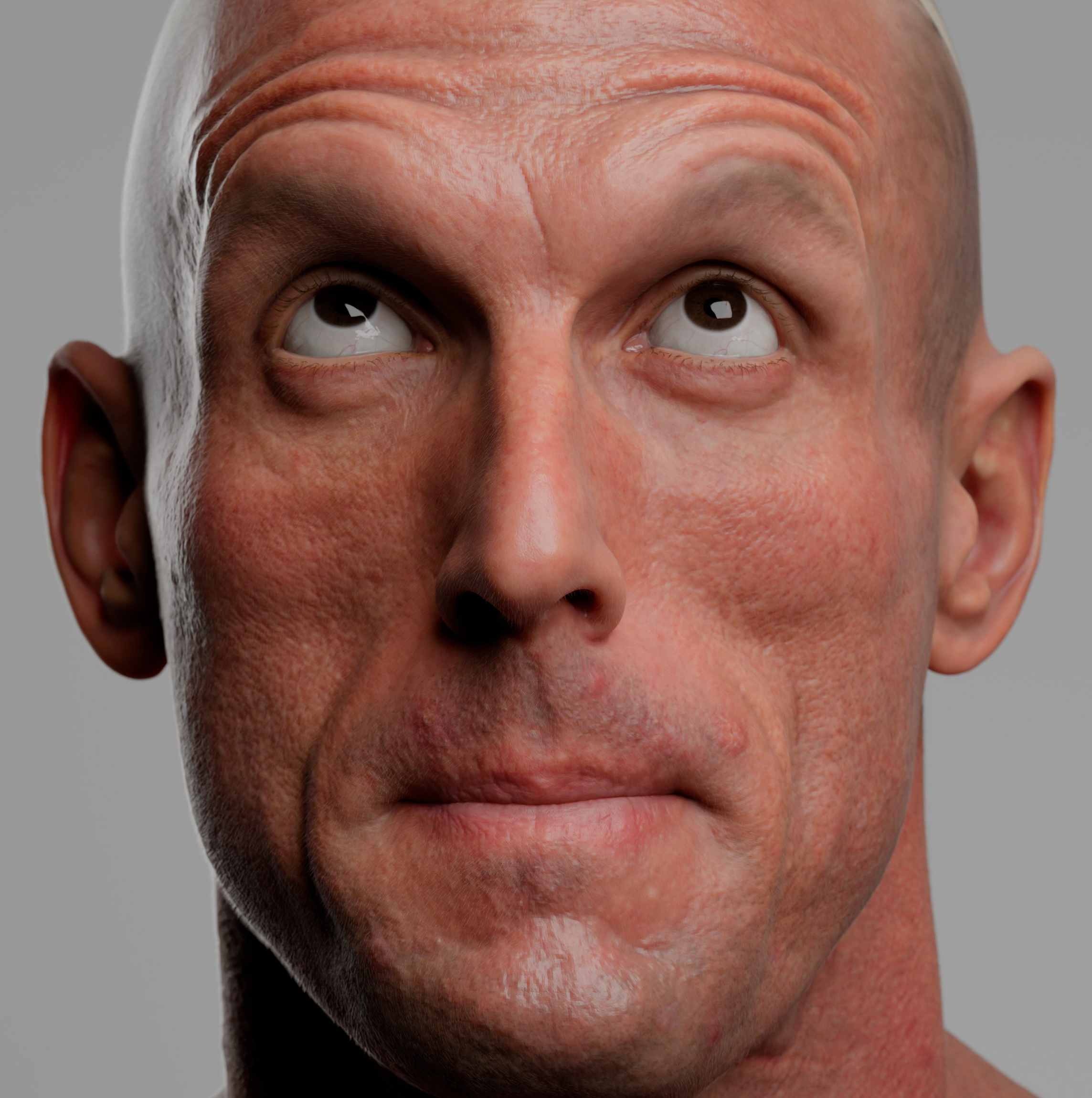} &
        \dreamfaceimg{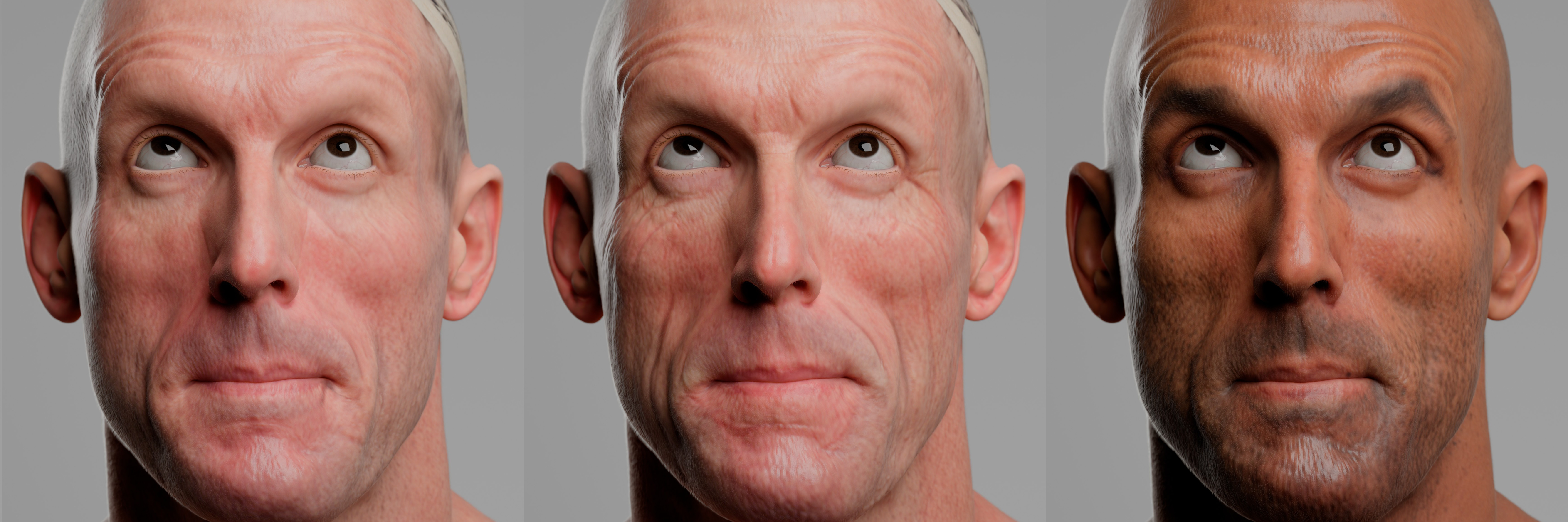} &
        \dreamfaceimg{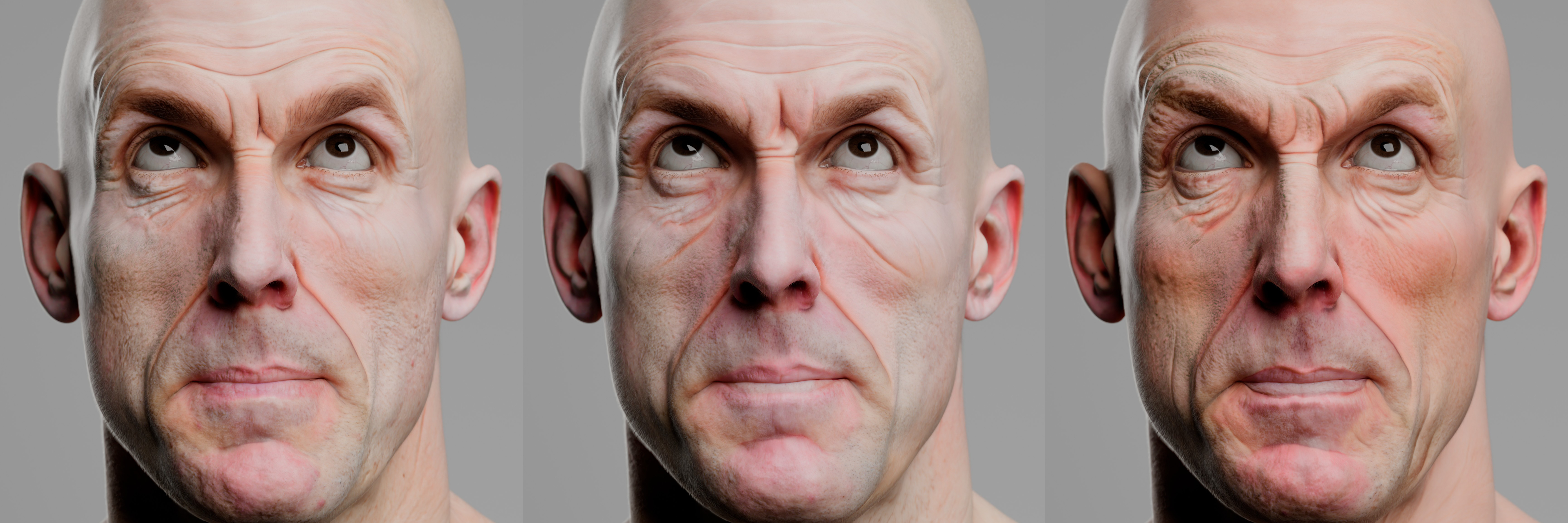} \\[0.1cm]
        \small \textbf{GT} & \small \textbf{FaceRiff (no-normal variant)} & \small \textbf{DreamFace}
    \end{tabular}
    \vspace{0.1cm}

    \textbf{Prompt 2:} \\
    \small\textit{``A < 24 African female with light brown skin and a slim build. The expression is neutral, suggesting a state of pensive. Facial analysis: Brows and forehead are in a relaxed, neutral position., Eyes are open with a direct gaze; no significant tension around the eyes or raising of the cheeks., Mouth is closed with lips lightly pressed together in a neutral line., Head is held straight and facing forward, with eyes looking directly at the camera.. The skin has an even, light brown tone with a smooth texture. There are a few small, hyperpigmented macules on the cheeks and around the nose. The skin appears well-hydrated with no visible wrinkles.''} \\[0.2cm]
    \begin{tabular}{@{}c@{\hspace{0.01\textwidth}}c@{\hspace{0.01\textwidth}}c@{}}
        \dreamfaceimg{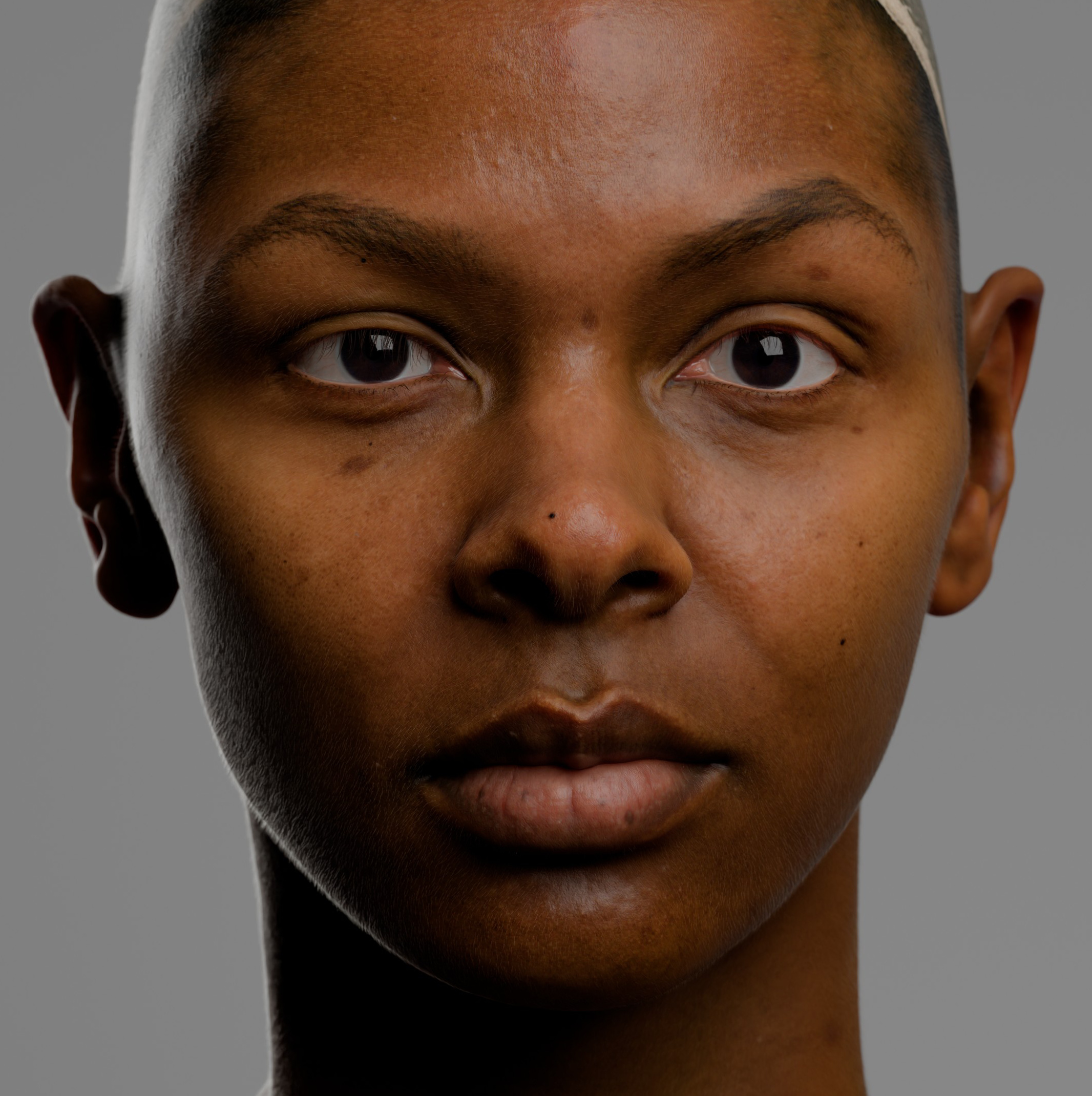} &
        \dreamfaceimg{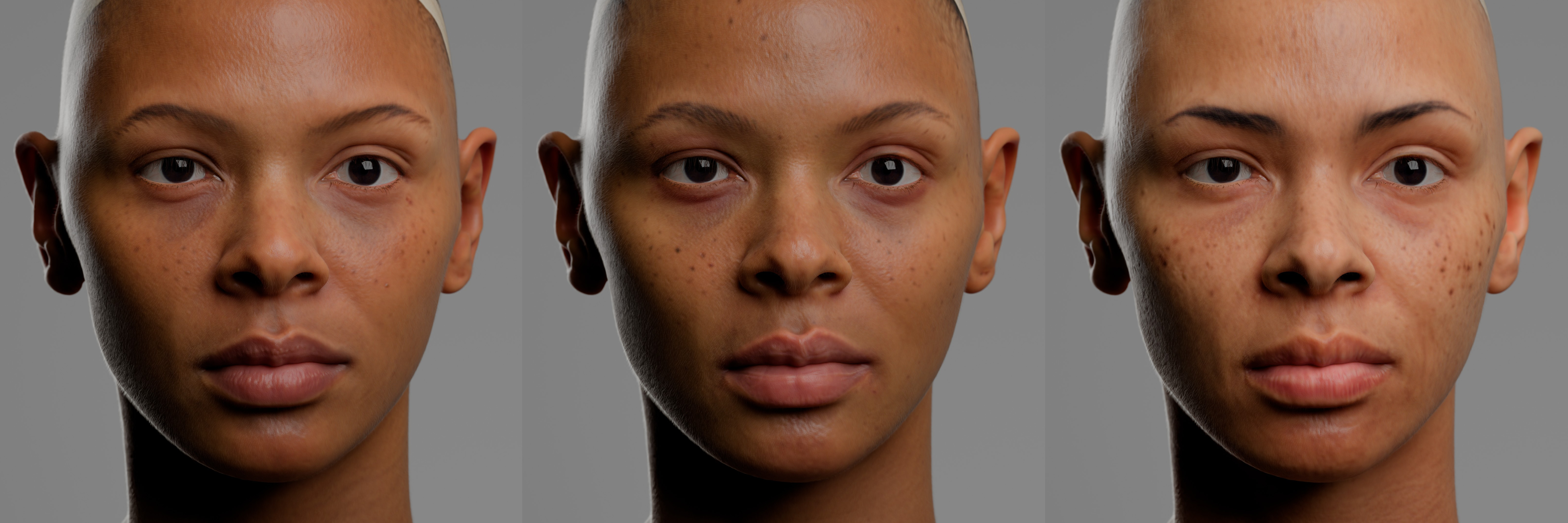} &
        \dreamfaceimg{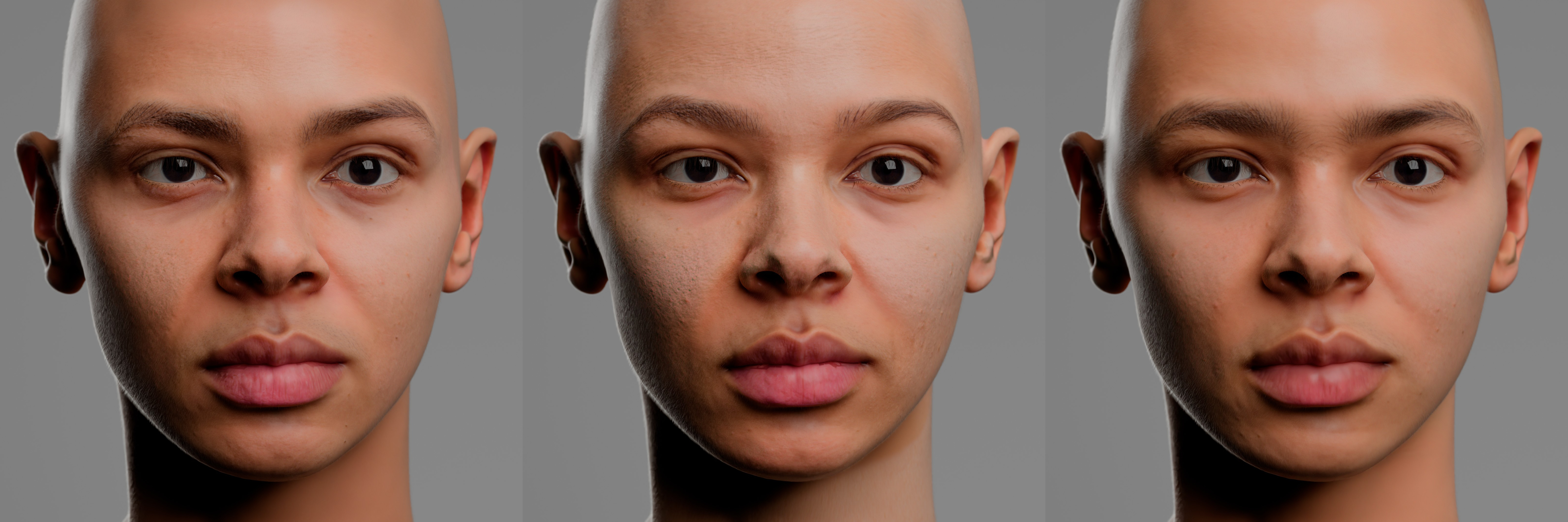} \\[0.1cm]
        \small \textbf{GT} & \small \textbf{FaceRiff (no-normal variant)} & \small \textbf{DreamFace}
    \end{tabular}
    \vspace{0.1cm}

    \textbf{Prompt 3:} \\
    \small\textit{``A 50+ Caucasian male with white skin and an average build. The expression is happy, suggesting a state of pleasant contemplation. Facial analysis: No significant brow muscle activation; static horizontal wrinkles are present., Slight activation of cheek raiser (AU 6) and lid tightener (AU 7), creating crow's feet., A subtle lip corner puller (AU 12) action creates a gentle, closed-mouth smile., Head is turned slightly to his left, eyes are gazing to the left.. The skin shows extensive signs of photoaging, with widespread freckles (lentigines) and uneven pigmentation on the forehead, cheeks, and nose. Static wrinkles are prominent around the eyes (crow's feet), and overall skin texture is slightly rough.''} \\[0.2cm]
    \begin{tabular}{@{}c@{\hspace{0.01\textwidth}}c@{\hspace{0.01\textwidth}}c@{}}
        \dreamfaceimg{images/dreamface_comp/gt_3495.jpg} &
        \dreamfaceimg{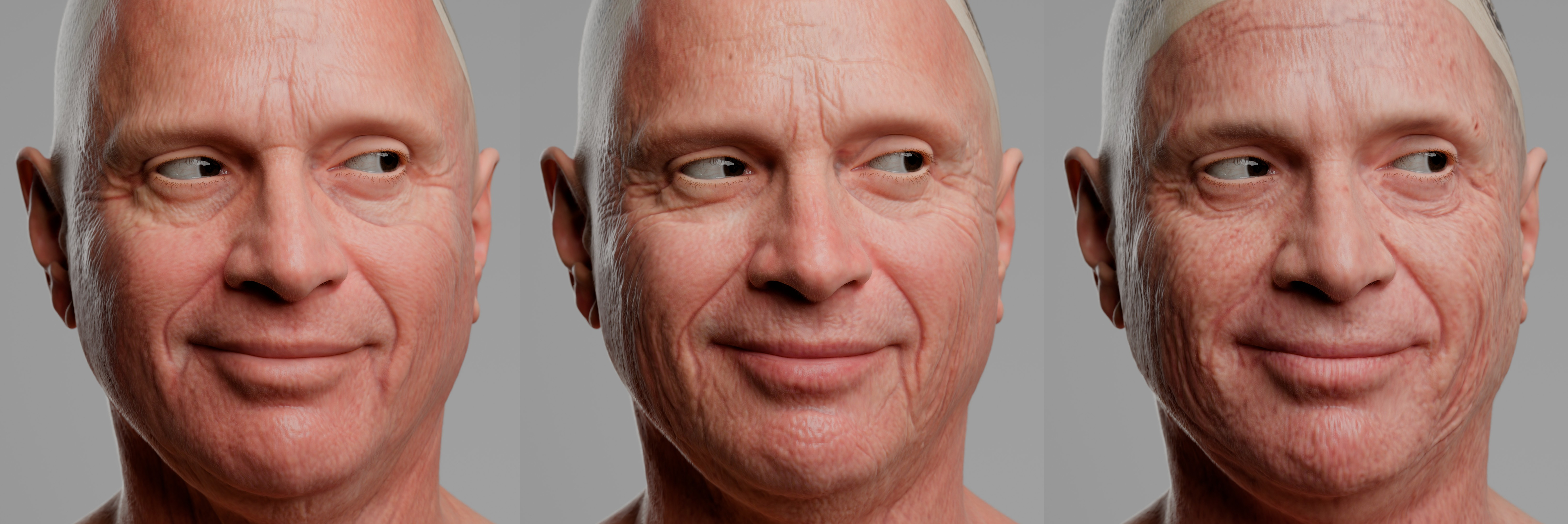} &
        \dreamfaceimg{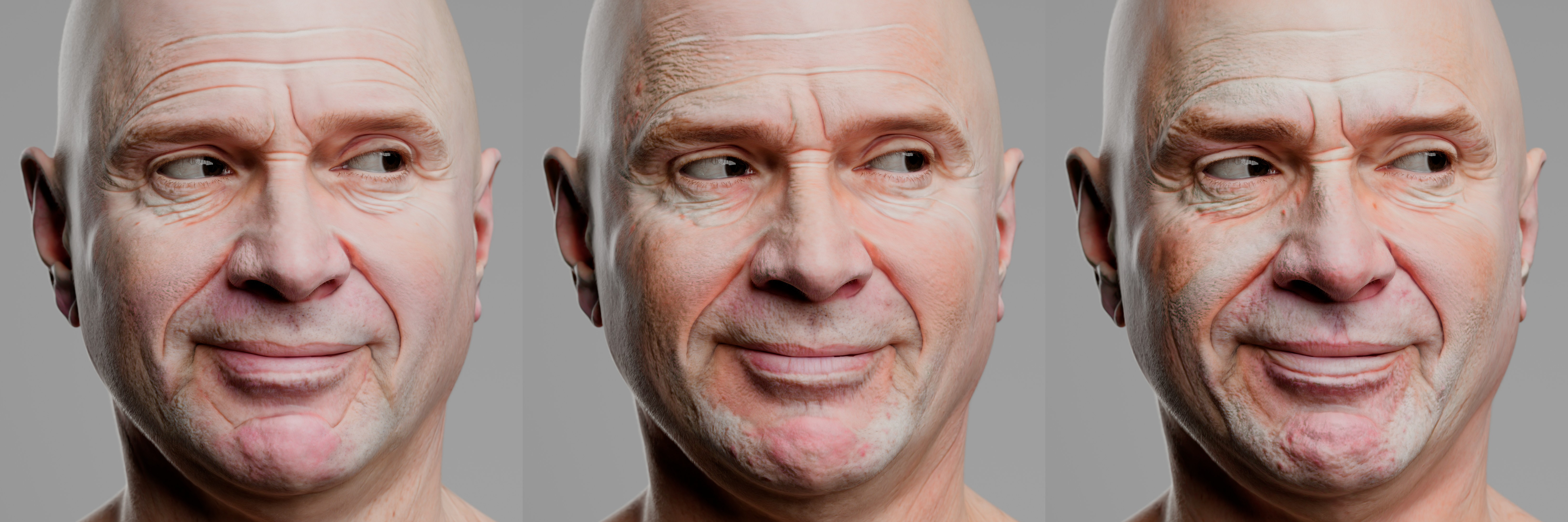} \\[0.1cm]
        \small \textbf{GT} & \small \textbf{FaceRiff (no-normal variant)} & \small \textbf{DreamFace}
    \end{tabular}
    \vspace{0.1cm}

    \textbf{Prompt 4:} \\
    \small\textit{``A < 24 Asian female with light brown skin and a slim build. The expression is neutral, suggesting a state of composed. Facial analysis: Brows and forehead are relaxed, no muscle activation detected., Eyelids are open in a neutral position with no squinting or cheek raising., Lips are closed and in a resting, neutral position., Head is facing forward and level, gaze is directed at the camera.. The skin has a generally smooth texture with visible pores on the cheeks and nose. There are a few scattered small nevi and minor areas of hyperpigmentation on the cheeks and chin.''} \\[0.2cm]
    \begin{tabular}{@{}c@{\hspace{0.01\textwidth}}c@{\hspace{0.01\textwidth}}c@{}}
        \dreamfaceimg{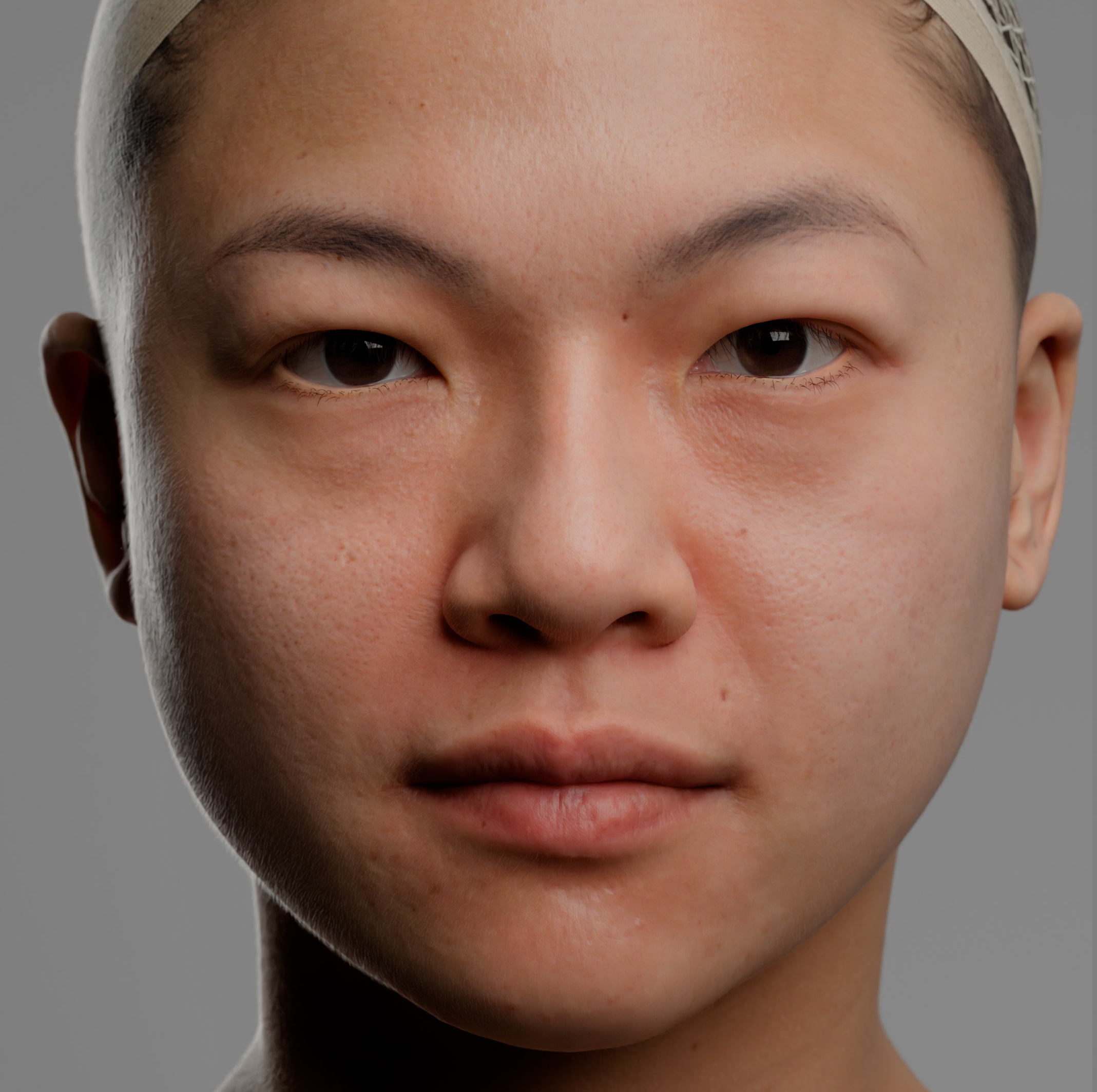} &
        \dreamfaceimg{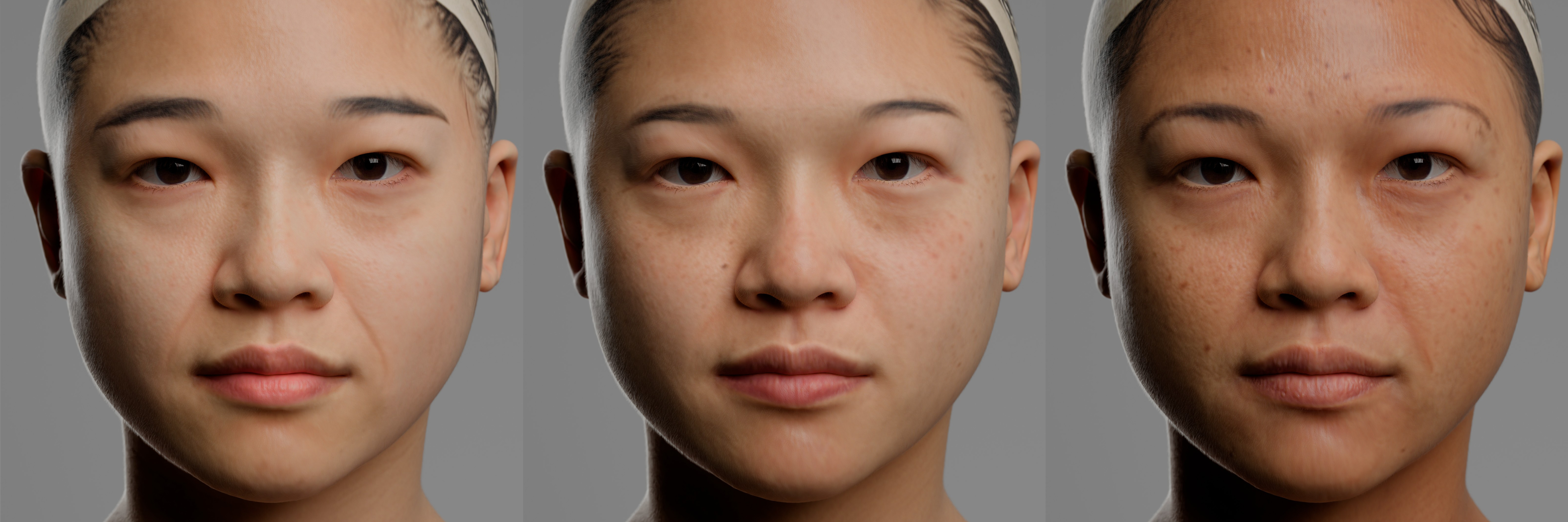} &
        \dreamfaceimg{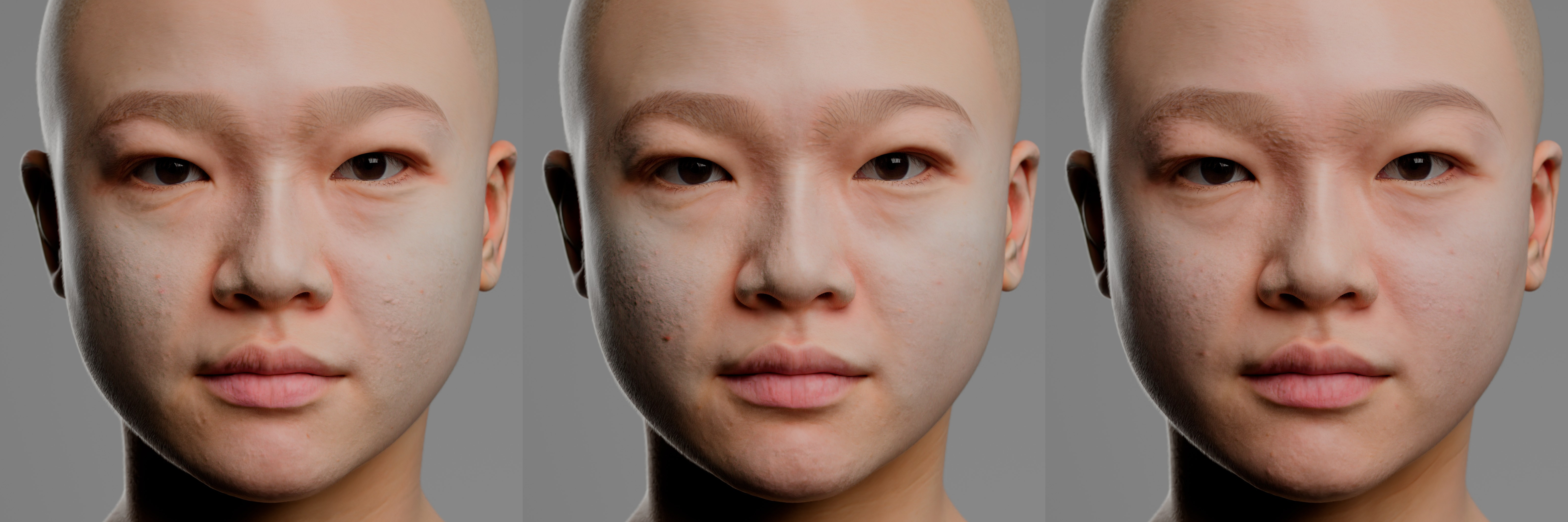} \\[0.1cm]
        \small \textbf{GT} & \small \textbf{FaceRiff (no-normal variant)} & \small \textbf{DreamFace}
    \end{tabular}

    \vspace{-0.3cm}
    \caption{\textbf{Fair comparison with DreamFace} (no-normal variant). For each prompt: GT (left), FaceRiff (middle), DreamFace (right). Across seeds, FaceRiff produces lighting-independent albedo with sharper pore detail and greater inter-sample diversity; DreamFace shows baked-in shading, smoother texture, and mode collapse across seeds.}
    \Description{Comparison of FaceRiff and DreamFace results for text-to-albedo generation, showing differences in lighting independence, pore-level detail, and diversity across four prompts.}
    \label{fig:dreamface_comparison}
\end{figure*}

\begin{figure*}[!p]
    \centering
    \begin{tabular}{c}
      \includegraphics[width=0.60\textwidth]{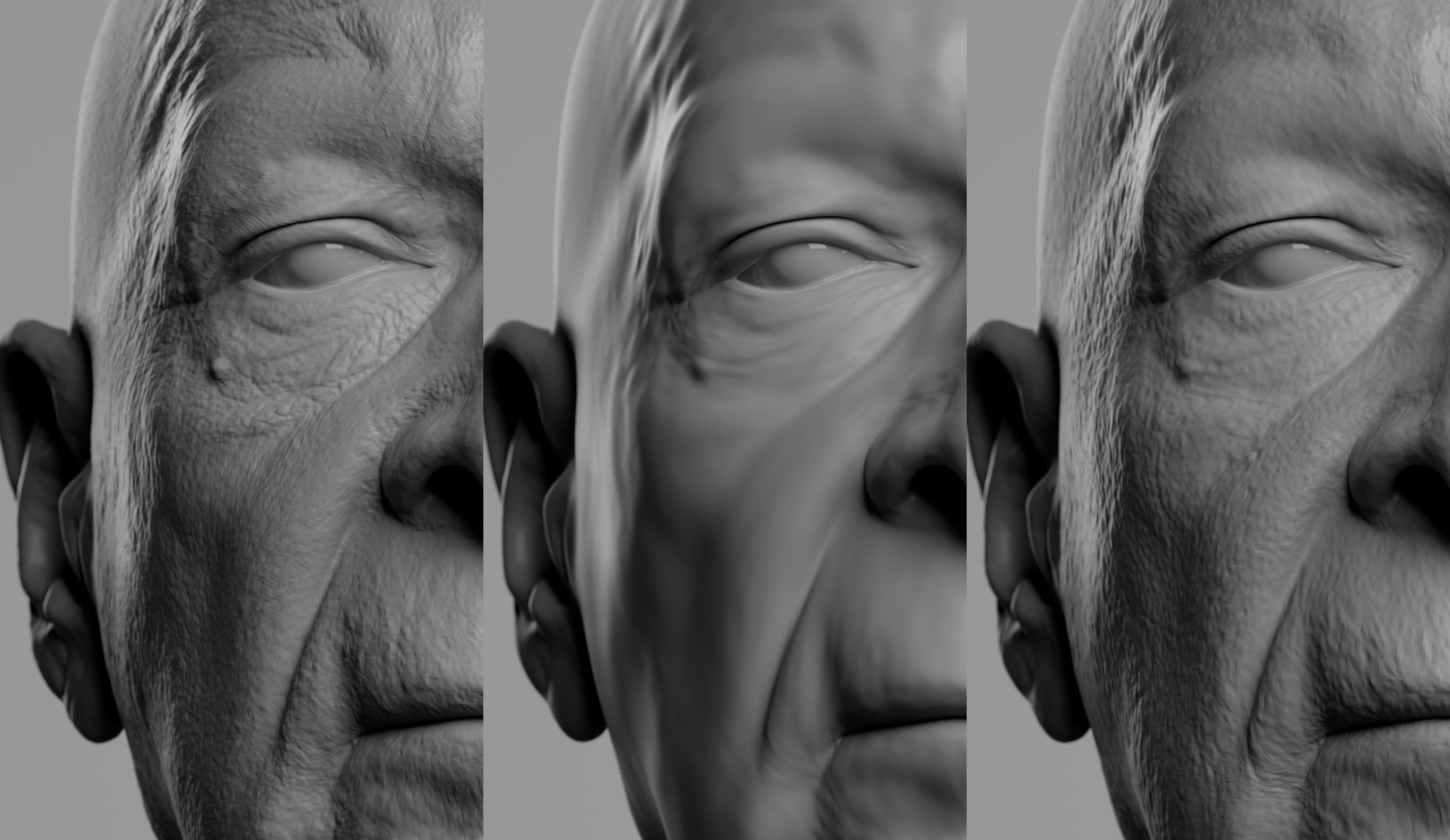} \\[0.1cm]
      \includegraphics[width=0.60\textwidth]{images/displacement/disp_fig_2406.jpg} \\[0.1cm]
      \includegraphics[width=0.60\textwidth]{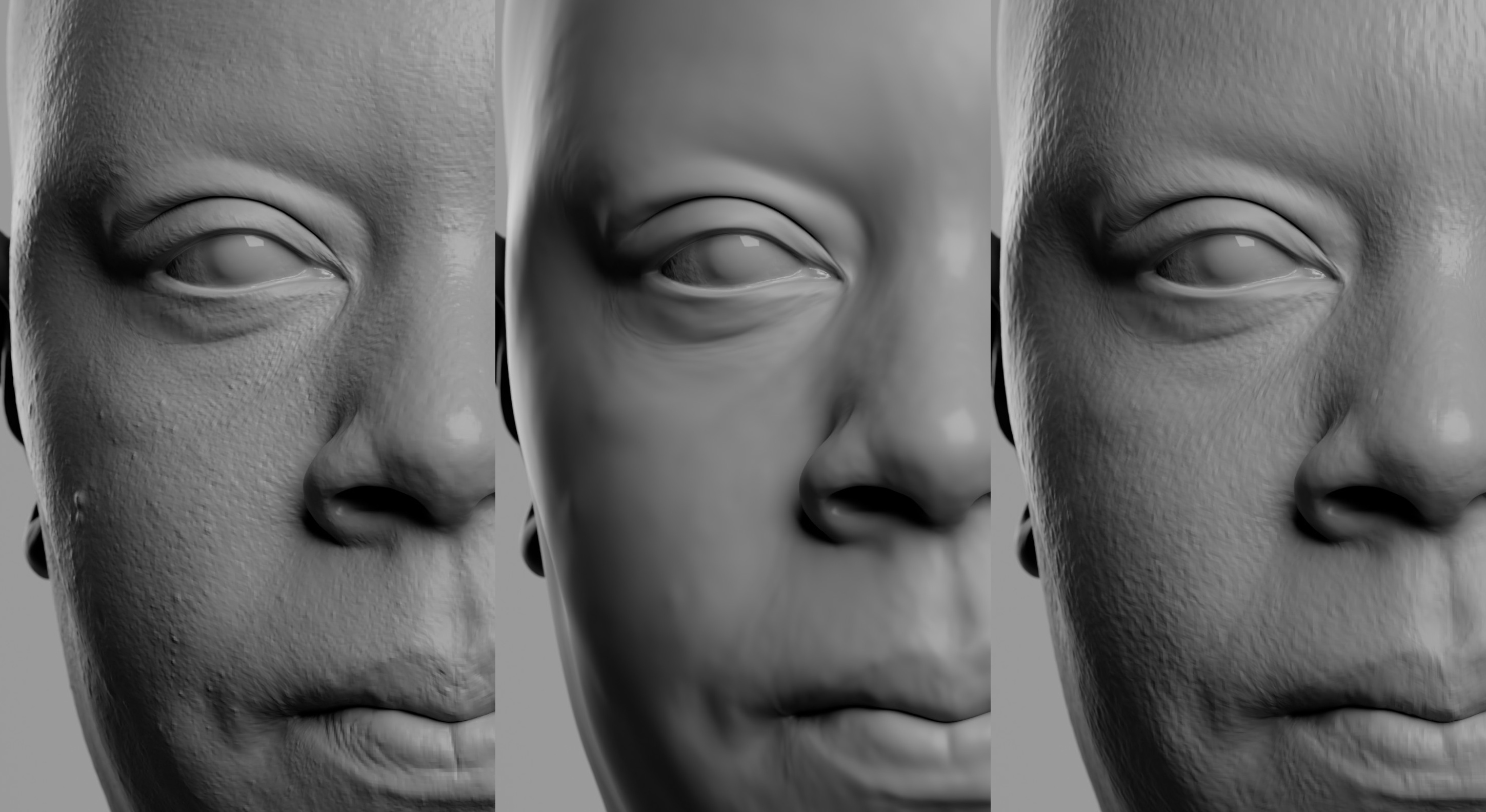}
    \end{tabular}
    \caption{\textbf{Additional text-to-displacement results.} Full-face comparisons (GT displacement, base mesh, FaceRiff) across diverse subjects, ages, and expressions. All results use normal conditioning with GT UV normals.}
    \Description{Multiple rows of text-to-displacement results showing ground-truth displacement, base mesh, and FaceRiff-generated displacement across diverse subjects with varying ages and expressions.}
    \label{fig:disp_supp}
\end{figure*}

\end{document}